\documentclass[]{bytedance_seed}
\usepackage{tabularx}

\usepackage[toc,page,header]{appendix}

\usepackage{minitoc}
\usepackage{cleveref} 
\usepackage{subcaption}
\usepackage{booktabs}
\usepackage{graphicx}
\usepackage{pgfplots}
\usepackage{pgfplotstable}
\usepackage{xcolor}
\usepackage{CJKutf8}
\usetikzlibrary{patterns}
\usepackage{float}
\usepackage{caption}
\newcommand{\method}{ByteWrist\xspace}

\usepackage{natbib}
\usepackage{latexsym}

\usepackage{url}
\usepackage{amssymb}
\usepackage[utf8]{inputenc}
\usepackage{microtype}
\usepackage{booktabs}
\usepackage{pifont} 
\usepackage{multirow}
\usepackage{makecell}
\usepackage{paralist}
\usepackage{xspace}
\usepackage{color}
\usepackage{xcolor}
\usepackage{colortbl}
\usepackage{adjustbox}
\usepackage{hyperref} 
\usepackage[edges]{forest}
\usepackage{tikz} 
\usepackage{caption}
\usepackage{amsfonts}

\hypersetup{
    colorlinks,
    linkcolor={blue!80!black},
    citecolor={blue!80!black},
}
\tikzset{
    root/.style =             {align=center, text width=1cm, rounded corners=3pt, line width=0.3mm, fill=gray!10, draw=gray!80, font=\small},
    demographic/.style =         {align=center, text width=1.8cm, rounded corners=3pt, line width=0.3mm, fill=blue!10, draw=blue!80, font=\footnotesize},
    demographic_work/.style =    {align=center, text width=10cm, rounded corners=3pt, line width=0.3mm, fill=blue!10, draw=blue!0, font=\footnotesize},
    character/.style =         {align=center, text width=1.8cm, rounded corners=3pt, line width=0.3mm, fill=red!10, draw=red!80, font=\footnotesize},
    character_work/.style =    {align=center, text width=10cm, rounded corners=3pt, line width=0.3mm, fill=red!10, draw=red!0, font=\footnotesize},
    personalization/.style =           {align=center, text width=1.8cm, rounded corners=3pt, line width=0.3mm, fill=cyan!10, draw=cyan!80, font=\footnotesize},
    personalization_work/.style =      {align=center, text width=10cm, rounded corners=3pt, line width=0.3mm, fill=cyan!10, draw=cyan!0, font=\footnotesize},
    risk/.style =         {align=center, text width=1.8cm, rounded corners=3pt, line width=0.3mm, fill=orange!10, draw=orange!80, font=\footnotesize},
    risk_work/.style =    {align=center, text width=10cm, rounded corners=3pt, line width=0.3mm, fill=orange!10, draw=orange!0, font=\footnotesize},
}

\usepackage{CJK}

\title{\method: A Parallel Robotic Wrist Enabling Flexible and Anthropomorphic Motion for Confined Spaces}

\author[\dagger]{Jiawen Tian}
\author[]{Liqun Huang}
\author[]{Zhongren Cui}
\author[]{Jingchao Qiao}
\author[]{Jiafeng Xu}
\author[]{Xiao Ma}
\author[\dagger]{Zeyu Ren}

\affiliation{ByteDance Seed}  

\contribution[\dagger]{Corresponding authors}

\date{September 23, 2025} 
\checkdata[Correspondence]{\url{tianjiawen.robot@bytedance.com,renzeyu.93@bytedance.com}}
\checkdata[Project Page]{\url{https://bytewrist.github.io/}}

\abstract{
This paper introduces ByteWrist, a novel highly-flexible and anthropomorphic parallel wrist for robotic manipulation. ByteWrist addresses the critical limitations of existing serial and parallel wrists in narrow-space operations through a compact three-stage parallel drive mechanism integrated with arc-shaped end linkages. The design achieves precise RPY (Roll-Pitch-Yaw) motion while maintaining exceptional compactness, making it particularly suitable for complex unstructured environments such as home services, medical assistance, and precision assembly. The key innovations include: (1) a nested three-stage motor-driven linkages that minimize volume while enabling independent multi-DOF control, (2) arc-shaped end linkages that optimize force transmission and expand motion range, and (3) a central supporting ball functioning as a spherical joint that enhances structural stiffness without compromising flexibility. Meanwhile, we present comprehensive kinematic modeling including forward / inverse kinematics and a numerical Jacobian solution for precise control. Empirically, we observe ByteWrist demonstrates strong performance in narrow-space maneuverability and dual-arm cooperative manipulation tasks, outperforming Kinova-based systems. Results indicate significant improvements in compactness, efficiency, and stiffness compared to traditional designs, establishing ByteWrist as a promising solution for next-generation robotic manipulation in constrained environments.
}

\begin{document}
\maketitle

\section{INTRODUCTION}
With the rapid advancement of robotic technology, robot application scenarios have shifted dramatically from open and highly structured environments to complex and unstructured settings such as home services, medical assistance, and precision assembly~\cite{1-unstructuredsettings, 2-unstructuredsettings}. Among these scenarios, operations in narrow and confined spaces, such as handling objects in cluttered home cabinets, performing minimally invasive surgical interventions in human body cavities, and assembling precision components in automotive engine compartments have put forward increasingly stringent requirements for the flexibility, compactness, and dynamic responsiveness of robotic end-effectors, especially robotic wrists.

Traditional serial robotic wrists, composed of multiple sequentially connected rotational joints, have the advantages of straightforward kinematic modeling~\cite{1-Traditionalserialroboticwrists,2-Traditionalserialroboticwrists} and a large rotational range per joint~\cite{3-Traditionalserialroboticwrists,4-Traditionalserialroboticwrists}. However, their open-chain structure leads to accumulated errors, low structural stiffness, and a relatively bulky overall volume. When operating in narrow spaces, the multi-link serial structure is prone to collision with the surrounding environment, limiting their maneuverability.
In contrast, parallel robotic wrists, which actuate the end platform through multiple parallel branches, demonstrate higher structural rigidity, superior load capacity, and better motion accuracy.

However, existing parallel wrist configurations with different structural designs and actuation methods struggle to simultaneously meet the requirements of compactness and flexibility~\cite{1-parallelroboticswrists}, as well as those of high rigidity, high load capacity and simple transmission mode~\cite{2-parallelroboticswrists, 3-parallelroboticswrists, 4-parallelroboticswrists, 5-parallelroboticswrists, 6-parallelroboticswrists, 7-parallelroboticswrists, 8-parallelroboticswrists}.
This renders them less suitable for anthropomorphic manipulation tasks that necessitate both high dexterity and strong spatial adaptability.

\begin{figure}[t]  
    \centering
    
    \begin{subfigure}[t]{0.46\columnwidth}  
        \centering
        \includegraphics[width=\linewidth]{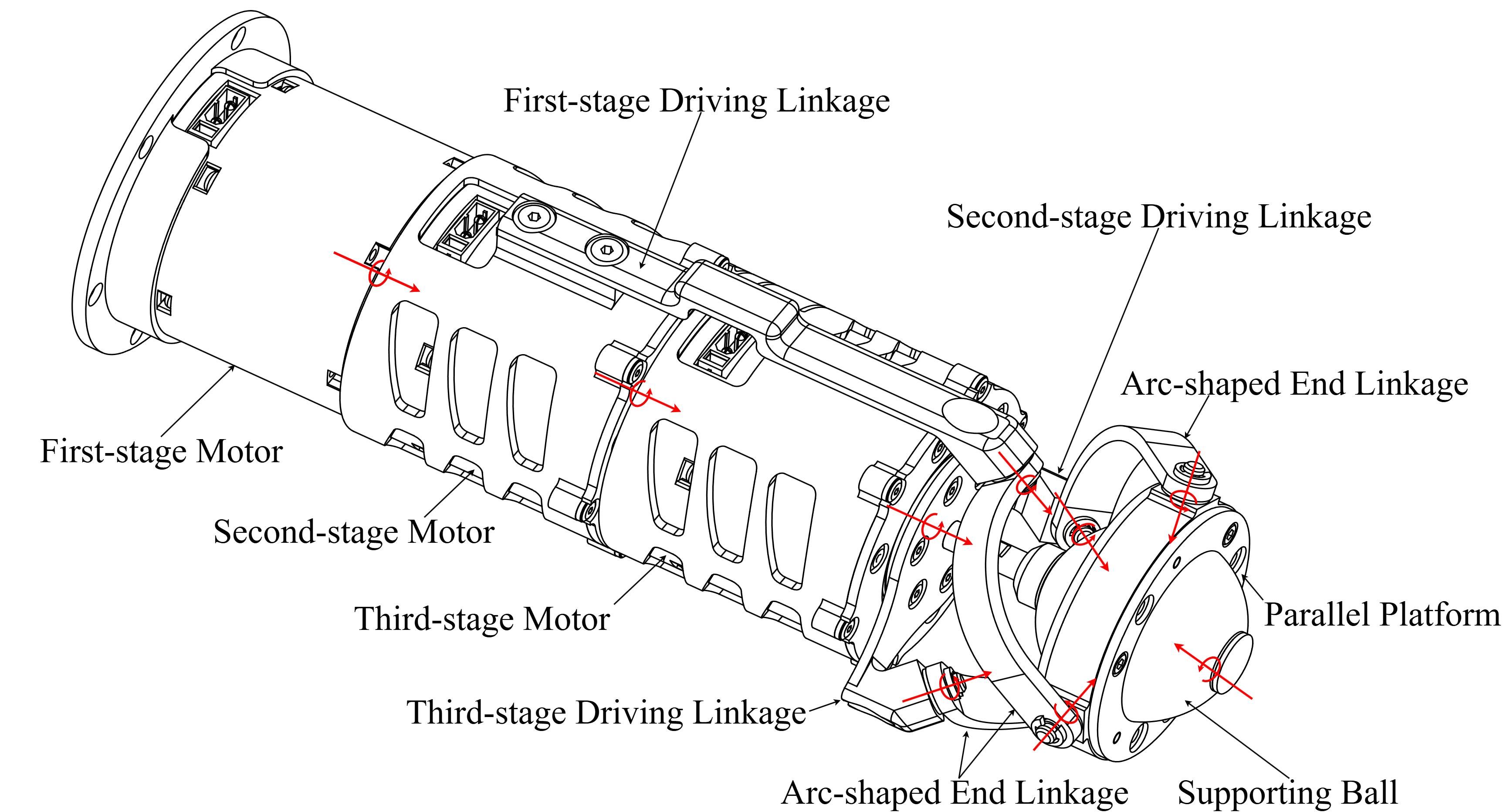}
        \caption{Components Description in Wireframe.}
        \label{fig:1-1}
    \end{subfigure}
    \hfill  
    \begin{subfigure}[t]{0.43\columnwidth}
        \centering
        \includegraphics[width=\linewidth]{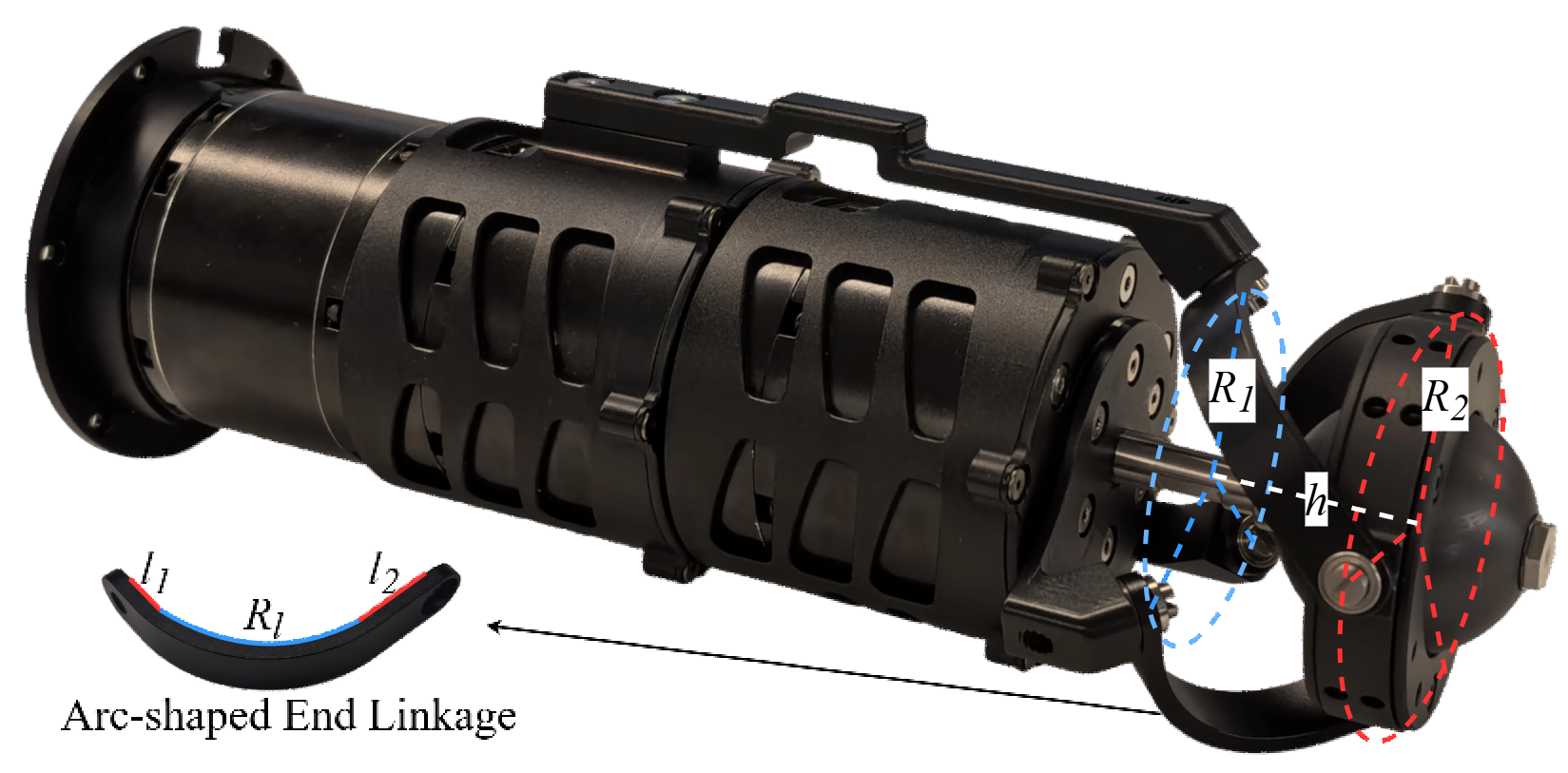}
        \caption{Parameters Description in Prototype.}
        \label{fig:1-2}
    \end{subfigure}
    
    \caption{Design and Development of ByteWrist.}
    \label{fig:1}
\end{figure}

To address these limitations, this paper proposes a novel parallel robotic wrist named ByteWrist. The core design concept of ByteWrist is to integrate a compact three-stage parallel drive mechanism with arc-shaped end linkages, achieving precise RPY motion \cite{1-RPY, 2-RPY, 3-RPY, 4-RPY} of the end platform while maintaining a small structural footprint. Compared with existing parallel and serial wrists, ByteWrist features three key innovations:

\begin{enumerate}
    \item \textbf{High Compactness:} The nested three-stage motor-driven linkages that minimize the overall volume while enabling independent control of multiple degrees of freedom. 
    \item \textbf{High Efficiency:} The arc-shaped end linkages optimize the force transmission path and expand the effective motion range of the end platform.
    \item \textbf{High Stiffness:} The central supporting ball functioned as a spherical joint enhances structural stiffness without sacrificing flexibility.
\end{enumerate}

This paper is organized as follows: Section II elaborates on the mechanical structure of ByteWrist, encompassing its drive mechanism, linkage design, and key structural parameters. Section III establishes the forward and inverse kinematic models for the wrist, and proposes a numerical solution for the Jacobian matrix to ensure precise control. Section IV validates the performance of ByteWrist through three sets of experiments: motion range testing, a narrow-space maneuverability comparison with Kinova, and a dual-arm chest-front cooperative manipulation for clothes hanging. Finally, Section V summarizes the research findings and discusses prospective improvements.

\section{STRUCTURE OF BYTEWRIST}

ByteWrist is illustrated in Fig. \ref{fig:1-1}, which is driven by three stage motors. The output of the first-stage motor is connected to the first-stage driving linkage, which is further linked to the parallel platform via an arc-shaped end linkage. Meanwhile, the second-stage motor is mounted inside the first-stage driving linkage, its output connects to the second-stage driving linkage, which is also linked to the parallel platform through an arc-shaped end linkage. Similarly, the third-stage motor is fixed within the second-stage driving linkage, with its output attached to the third-stage driving linkage that connects to the parallel platform via an arc-shaped end linkage.

All three stage driving linkages and arc-shaped end linkages, as well as arc-shaped end linkages and parallel platform, are connected via revolute pairs. All these six revolute pairs are oriented toward the center of the parallel platform.

To enhance the stiffness of the parallel platform, a supporting ball is mounted at its center and connected to the platform via a spherical joint. By controlling the movement of three stage motors, the end parallel platform can achieve precise RPY motion.

As illustrated in Fig. \ref{fig:1-2}, the prototype of ByteWrist adopts Quasi-Direct Drive \cite{QDD, QDD1, QDD2} based actuators manufactured by \href{https://robstride.com/}{RobStride Dynamics}. \(R_1\) denotes the rotation radius of the connection point between all three stage driving linkages and arc-shaped end linkages,  \(R_2\) denotes the rotation radius of the connection point between arc-shaped end linkages and the parallel platform. The center distance between these two rotation planes is \(h\).

The arc-shaped end linkage consists of a 90-degree arc and two straight segments connected to the arc, where the radius of the arc is \(R_l\) and the length of two straight segments are \(l_1\) and \(l_2\) respectively, as described in Fig. \ref{fig:1-2}.

Above mentioned parameters are shown in Table \ref{table_1}.

\begin{table}[h]
    \centering
    \caption{Structural Parameters of ByteWrist (Unit:\,mm)}
    \label{table_1}
    \begin{tabular}{cccccc}
        \toprule
        \(R_1\) & \(R_2\) & \(h\) & \(R_l\) & \(l_1\) & \(l_2\) \\
        \midrule
        27.35 & 30 & 27.35 & 25 & 5 & 13.68 \\
        \bottomrule
    \end{tabular}
\end{table}

\section{FORWARD AND INVERSE KINEMATICS OF BYTEWRIST}
\begin{figure}[htbp]
    \centering
    \includegraphics[width=0.7\linewidth]{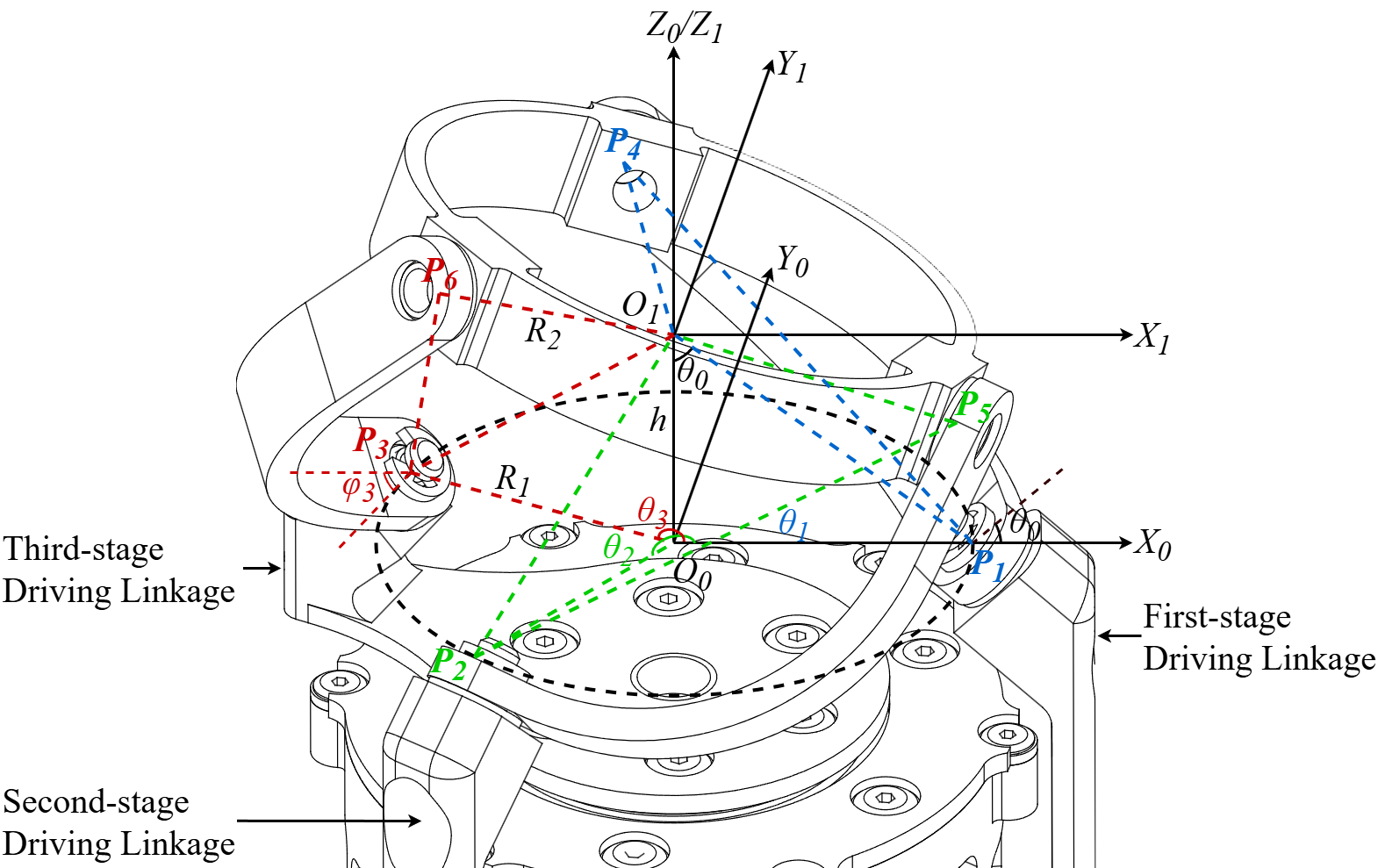}
    \caption{Coordinate System Definition of ByteWrist.}
    \label{fig:2}
\end{figure}

To achieve precise control of ByteWrist, deriving its forward and inverse kinematics is essential. For clarity of description, its structure is simplified as illustrated in Fig. \ref{fig:2}.

Points \(P_1\), \(P_2\) and \(P_3\) lie on the axis of the revolute joint connecting the arc-shaped end linkage and the driving linkage, and are situated on the inner surface of arc-shaped end linkages. Points \(P_4\), \(P_5\), and \(P_6\) lie on the axis of the revolute joint between the arc-shaped end linkage and the parallel platform, and are similarly located on the inner surface of arc-shaped end linkages. Point \(O_0\) is the center of the circle passing through points \(P_1\), \(P_2\) and \(P_3\) with a radius of \(R_1\). Correspondingly point \(O_1\) is the center of the circle passing through points \(P_4\), \(P_5\) and \(P_6\) with a radius of \(R_2\). For the coordinate system \(\left\{ O_0X_0Y_0Z_0\right\}\), when the parallel wrist is in its initial position, \(\vec{O_0X_0}\) points to \(P_1\), while \(\vec{O_0Z_0}\) aligns with the axis of the driving motor, with its positive direction defined as Fig. \ref{fig:2} presents. For the coordinate system \(\left\{ O_1X_1Y_1Z_1\right\}\), its coordinate axis are parallel to coordinate system \(\left\{ O_0X_0Y_0Z_0\right\}\), and the distance between \(O_1\) and \(O_0\) is \(h\). \(\theta_0\) is the angle of \(\angle  P_1O_1O_0\), and \(\theta_1\), \(\theta_2\), \(\theta_3\) are respectively the angles between the three driving linkages and the \(\vec{O_0X_0}\) direction, note that at initial position \(\theta_1 = 0^\circ\), \(\theta_2 = 240^\circ\), \(\theta_3 = 120^\circ\).

Generally, in the design process, \(R_1\), \(R_2\), and \(h\) are structural parameters to be determined with high priority by taking the wrist actuators diameter into consideration. Once these three parameters are confirmed, the relationships between the parameters \(R_l\), \(l_1\), and \(l_2\) of the arc-shaped end linkage can be calculated using (1). After \(R_l\) is further determined, the overall structural design can be completed.
\begin{equation}
\left\{\begin{matrix}
\theta_0=\arctan (R_1/h)\\
R_l+l_2=h/\cos\theta _0\\
R_l+l_1=R_2\end{matrix}\right.
\end{equation}

\subsection{Forward Kinematics}
The kinematics of the parallel wrist involves solving the relationship between \(\theta_1\), \(\theta_2\), \(\theta_3\) and the RPY angles of the parallel platform. For the forward kinematics part, the inputs are \(\theta_1\), \(\theta_2\), \(\theta_3\), and the outputs are RPY angles of the parallel platform.

When three driving linkages rotate to \(\theta_1\), \(\theta_2\), and \(\theta_3\), the arc-shaped end linkages \(P_1P_4\), \(P_2P_5\) and \(P_3P_6\) rotate around \(\vec{O_1P_1}\), \(\vec{O_1P_2}\) and \(\vec{O_1P_3}\) by \(\varphi_1\), \(\varphi_2\) and \(\varphi_3\) respectively. As can be seen from the geometric structure of the parallel wrist, the coordinates of points \(P_1\), \(P_2\), and \(P_3\) are \(P_i=(R_1\cos\theta_i, R_1\sin\theta_i, 0)\)\((i=1,2,3)\). 

The coordinates of points \(P_i\)\((i=1\sim6)\) in their initial positions are denoted as \(P_{i0}\). In the initial position, \(\theta_1=0\), \(\theta_2=4\pi/3\), and \(\theta_3=2\pi/3\). It can be calculated that \(P_{40}=(-R_2\sin\theta_1,R_2\cos\theta_1,h)\), \(P_{50}=(-R_2\sin\theta_2,R_2\cos\theta_2,h)\), \(P_{60}=(-R_2\sin\theta_3,R_2\cos\theta_3,h)\).

When the parallel wrist moves to any position, the coordinates of points \(P_4\), \(P_5\), and \(P_6\) can be derived as (2).
\begin{equation}
\left\{\begin{matrix}
P_{i+3}.x= -R_2\cos\varphi_i\sin\theta_i+R_2\sin\varphi_i\cos\theta_0\cos\theta_i \\
P_{i+3}.y= R_2\cos\varphi_i\cos\theta_i+R_2\sin\varphi_i\cos\theta_0\sin\theta_i \\
P_{i+3}.z= R_2\sin\varphi_i\sin\theta_0+h \\
(i=1,2,3)\end{matrix}\right.
\end{equation}

Thus vectors \(\vec{O_1P_4}\), \(\vec{O_1P_5}\), \(\vec{O_1P_6}\) can be expressed in (3).
\begin{equation}
\vec{O_1P_{i+3}}=(P_{i+3}.x,P_{i+3}.y,R_2\sin\varphi_i\sin\theta_0) (i=1,2,3)
\end{equation}

The angle between each pair of these three vectors is \(2\pi/3\), and (4) can be obtained.
\begin{equation}
\left\{\begin{matrix}
\frac{\vec{O_1P_4}\cdot\vec{O_1P_5}}{\left |\vec{O_1P_4} \right |\cdot\left |\vec{O_1P_5} \right |}=\cos\frac{2\pi}{3}\\
\frac{\vec{O_1P_4}\cdot\vec{O_1P_6}}{\left |\vec{O_1P_6} \right |\cdot\left |\vec{O_1P_6} \right |}=\cos\frac{2\pi}{3}\\
\frac{\vec{O_1P_5}\cdot\vec{O_1P_6}}{\left |\vec{O_1P_5} \right |\cdot\left |\vec{O_1P_6} \right |}=\cos\frac{2\pi}{3}\end{matrix}\right.
\end{equation}

By solving the system of equations (4) simultaneously, the values of \(\varphi_1\), \(\varphi_2\) and \(\varphi_3\) can be calculated for any given inputs \(\theta_1\), \(\theta_2\), \(\theta_3\). Given that this system of equations is nonlinear, the Newton-Raphson method is adopted herein for iterative solution.

This paper adopts the RPY method to describe the attitude of the parallel platform. Let \(\left\{ O_2X_2Y_2Z_2\right\}\) denote the coordinate system of the parallel platform after rotation. Specifically, the platform undergoes rotations around the X-axis, Y-axis and Z-axis of the original coordinate system \(\left\{ O_1X_1Y_1Z_1\right\}\) by \(\gamma\)(Roll), \(\beta\)(Pitch) and \(\alpha\)(Yaw) respectively. The rotation matrix describing the attitude transformation of the parallel platform is expressed in (5).

\begin{equation}
\begin{aligned}
R_X(\gamma) &= \begin{bmatrix}
1 & 0 & 0 \\
0 & \cos\gamma & -\sin\gamma \\
0 & \sin\gamma & \cos\gamma
\end{bmatrix}, \\
R_Y(\beta) &= \begin{bmatrix}
\cos\beta & 0 & \sin\beta \\
0 & 1 & 0 \\
-\sin\beta & 0 & \cos\beta
\end{bmatrix}, \\
R_Z(\alpha) &= \begin{bmatrix}
\cos\alpha & -\sin\alpha & 0 \\
\sin\alpha & \cos\alpha & 0 \\
0 & 0 & 1
\end{bmatrix}, \\
^1R_2 &= R_Z(\alpha)R_Y(\beta)R_X(\gamma) \\
&= \begin{bmatrix}
R_{11} & R_{12} & R_{13} \\
R_{21} & R_{22} & R_{23} \\
R_{31} & R_{32} & R_{33}
\end{bmatrix}
\end{aligned}
\end{equation}

The normal vector of the parallel platform is \(n = (A, B, C)\), where \(A\), \(B\), and \(C\) can be calculated by (6):
\begin{equation}
\begin{cases}
A = (P_{5}.y - P_{4}.y)(P_{6}.z - P_{4}.z) - (P_{5}.z - P_{4}.z)(P_{6}.y - P_{4}.y) \\
B = (P_{5}.z - P_{4}.z)(P_{6}.x - P_{4}.x) - (P_{5}.x - P_{4}.x)(P_{6}.z - P_{4}.z) \\
C = (P_{5}.x - P_{4}.x)(P_{6}.y - P_{4}.y) - (P_{5}.y - P_{4}.y)(P_{6}.x - P_{4}.x)
\end{cases}
\end{equation}

\(R_{12}\), \(R_{22}\), and \(R_{32}\) can be calculated by (7).
\begin{equation}
\begin{bmatrix}
P_4.x \\
P_4.y \\
P_4.z - h
\end{bmatrix}
=
\,^1R_2 
\begin{bmatrix}
P_{40}.x \\
P_{40}.y \\
P_{40}.z-h
\end{bmatrix}
=
\,^1R_2 
\begin{bmatrix}
0 \\
R_2 \\
0
\end{bmatrix}
\end{equation}

Furthermore, \(R_{11}\), \(R_{21}\), and \(R_{31}\) can be calculated by (8).
\begin{equation}
\begin{bmatrix}
P_5.x \\
P_5.y \\
P_5.z - h
\end{bmatrix}
=
\,^1R_2 
\begin{bmatrix}
P_{50}.x \\
P_{50}.y \\
P_{50}.z-h
\end{bmatrix}
=
\,^1R_2 
\begin{bmatrix}
-R_2\sin\frac{4\pi}{3} \\
R_2\cos\frac{4\pi}{3} \\
0
\end{bmatrix}
\end{equation}

\(R_{13}\), \(R_{23}\), and \(R_{33}\) can be calculated by (9).
\begin{equation}
\begin{bmatrix}
A \\
B \\
C
\end{bmatrix}
= \,^1R_2 
\begin{bmatrix}
0 \\
0 \\
\sqrt{A^2 + B^2 + C^2}
\end{bmatrix}
\end{equation}

Then, the RPY angles of the parallel platform can be solved by (10):
\begin{equation}
\left\{\begin{matrix}
\alpha=atan2(R_{21},R_{11})\\
\beta=atan2(-R_{31},\sqrt{R_{32}^2+R_{33}^2})\\
\gamma=atan2(R_{32},R_{33})\end{matrix}\right.
\end{equation}

\subsection{Inverse Kinematics}
For inverse kinematics, the input is RPY angles of the parallel platform, and the output is \(\theta_1\), \(\theta_2\) and \(\theta_3\). After given the RPY angles, the rotation matrix \(^1R_2\) can be solved through (5). Furthermore, the coordinates of points \(P_4\), \(P_5\), and \(P_6\) can be calculated using (7), (8), and (11).
\begin{equation}
\begin{bmatrix}
P_6.x \\
P_6.y \\
P_6.z - h
\end{bmatrix}
=
\,^1R_2 
\begin{bmatrix}
P_{60}.x \\
P_{60}.y \\
P_{60}.z-h
\end{bmatrix}
=
\,^1R_2 
\begin{bmatrix}
-R_2\sin\frac{2\pi}{3} \\
R_2\cos\frac{2\pi}{3} \\
0
\end{bmatrix}
\end{equation}

The values of \(\varphi_1\), \(\varphi_2\), \(\varphi_3\), \(\theta_1\), \(\theta_2\), and \(\theta_3\) can be derived using (2).

\subsection{Jacobian Matrix}
This paper adopts the numerical method to solve the Jacobian matrix, which can be expressed as (12):
\begin{equation}
J = \begin{bmatrix}
J_{11} & J_{12} & J_{13} \\
J_{21} & J_{22} & J_{23} \\
J_{31} & J_{32} & J_{33} \\
\end{bmatrix} 
= \begin{bmatrix}
\frac{\partial \alpha}{\partial \theta_1} & \frac{\partial \alpha}{\partial \theta_2} & \frac{\partial \alpha}{\partial \theta_3} \\
\frac{\partial \beta}{\partial \theta_1} & \frac{\partial \beta}{\partial \theta_2} & \frac{\partial \beta}{\partial \theta_3} \\
\frac{\partial \gamma}{\partial \theta_1} & \frac{\partial \gamma}{\partial \theta_2} & \frac{\partial \gamma}{\partial \theta_3}
\end{bmatrix}
\end{equation}

For a small step \(\Delta\theta\), \(J_{11} \) can be calculated by (13).
\begin{equation}
J_{11}=\frac{\partial \alpha}{\partial \theta_1}\approx \frac{\alpha(\theta_1+\Delta \theta,\theta_2,\theta_3)}{\Delta \theta}
\end{equation}

The other elements of the Jacobian matrix \(J\) can be calculated in the same way.

As discussed previously, the iterative Newton-Raphson method inherently gives rise to rounding errors. In the numerical solution of the Jacobian matrix of the wrist, special consideration must be given to the effect of step size \(\Delta\theta\) on the precision of the results.


The motion process of the wrist under the conditions \(\theta_1=0,\theta_2=4\pi/3\) and \(\theta_3\) ranging from \(2\pi/3\) to \(\pi\) is simulated using Creo software. The RPY angles of the parallel platform were recorded at a step size of \(\pi/3000\). A series of step sizes (\(1e-1\), \(1e-2\), ..., \(1e-7\)) are selected to calculate both the maximum error and root mean square error (RMSE) of the elements in Jacobian matrix.  Comparative analysis revealed that the smallest values of both maximum error and RMSE for matrix J occur when the step size \(\Delta\theta\) is \(1e-3\). Consequently, a step size of \(\Delta\theta\) = \(1e-3\) is chosen for solving the Jacobian matrix via the numerical method.

\section{EXPERIMENTS}
In this chapter, three experiments are conducted to verify the performance of ByteWrist in terms of flexibility, high integration and anthropomorphic. Prior to these experiments, an introduction to the ByteMini robotic system is provided.

\subsection{Prototype of ByteMini}

\begin{figure}[htbp]
    \centering
    \includegraphics[width=0.65\linewidth]{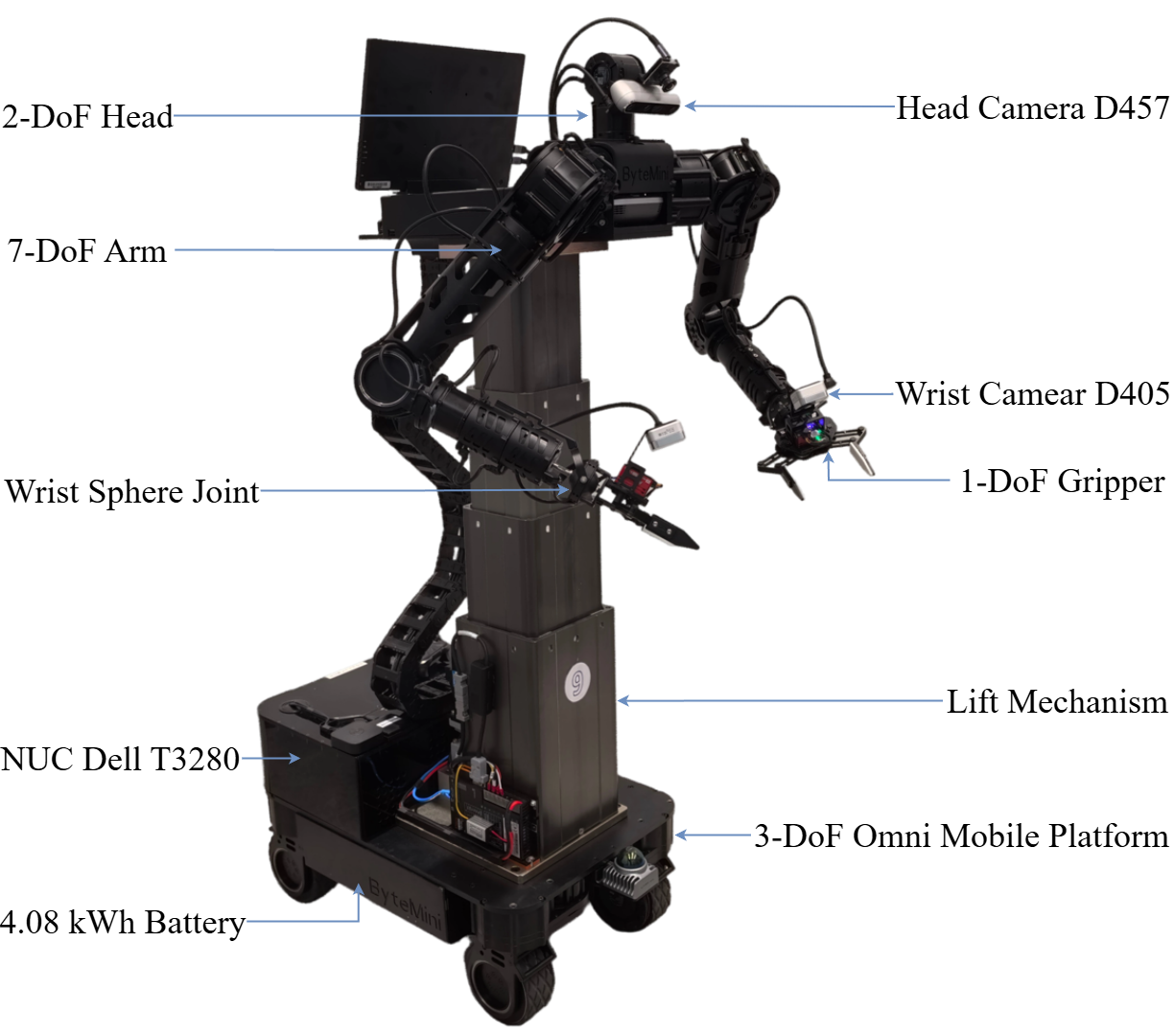}
    \caption{Description of the ByteMini Robot.}
    \label{fig:3}
\end{figure}

To verify ByteWrist functionality, we integrate them into our 22-DoF dual-arm mobile robot, ByteMini (Fig. \ref{fig:3}). Key design features include:

\begin{enumerate}
    \item \textbf{Arms:}\,7-DoF in SRS (Spherical-Revolute-Spherical) configuration with ByteWrist as wrist modules.
    \item \textbf{Grippers:}\,1-DoF grippers paired with RealSense D405 cameras for close-range vision.
    \item \textbf{Waist:}\,1-DoF lifting mechanism in high stiffness for height adjustment.
    \item \textbf{Chassis:}\,3-DoF omnidirectional mobile platform for flexible movement.
    \item \textbf{Head:}\,2-DoF in Pitch and Yaw motion, integrated with a RealSense D457 camera for primary vision.
    \item \textbf{Computing and Power:}\,Dell T3280 computer as the main controller and 4.08 kWh battery for power supply.
 
\end{enumerate}

\subsection{Wrist Motion Range}
Owing to structural constraints, the motion range of ByteWrist must comply with the requirement specified in (14).
\begin{equation}
\beta^2+\gamma^2<0.7^2
\end{equation}

To verify the wrist range motion and high flexibility, this study designs a circular trajectory, enabling the wrist to move in accordance with (15).
\begin{equation}
\left\{\begin{matrix}
\beta^2+\gamma^2=0.68\\
\alpha=0\end{matrix}\right.
\end{equation}

\begin{figure*}[t]
    \vspace{5pt}
    \centering
    \setlength{\tabcolsep}{4pt} 

    \begin{subfigure}[t]{0.32\textwidth}  
        \centering
        \includegraphics[width=\textwidth]{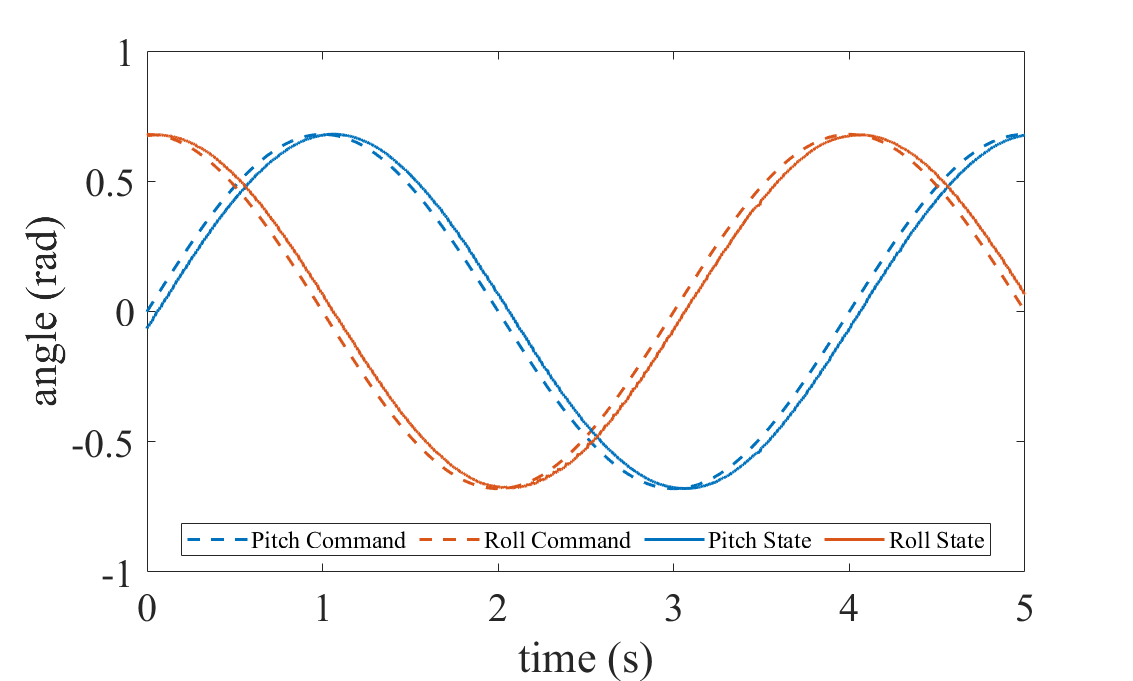}
        \caption{T = 4 s}
        \label{fig:4-9}
    \end{subfigure}
    \hfill
    \begin{subfigure}[t]{0.32\textwidth}
        \centering
        \includegraphics[width=\textwidth]{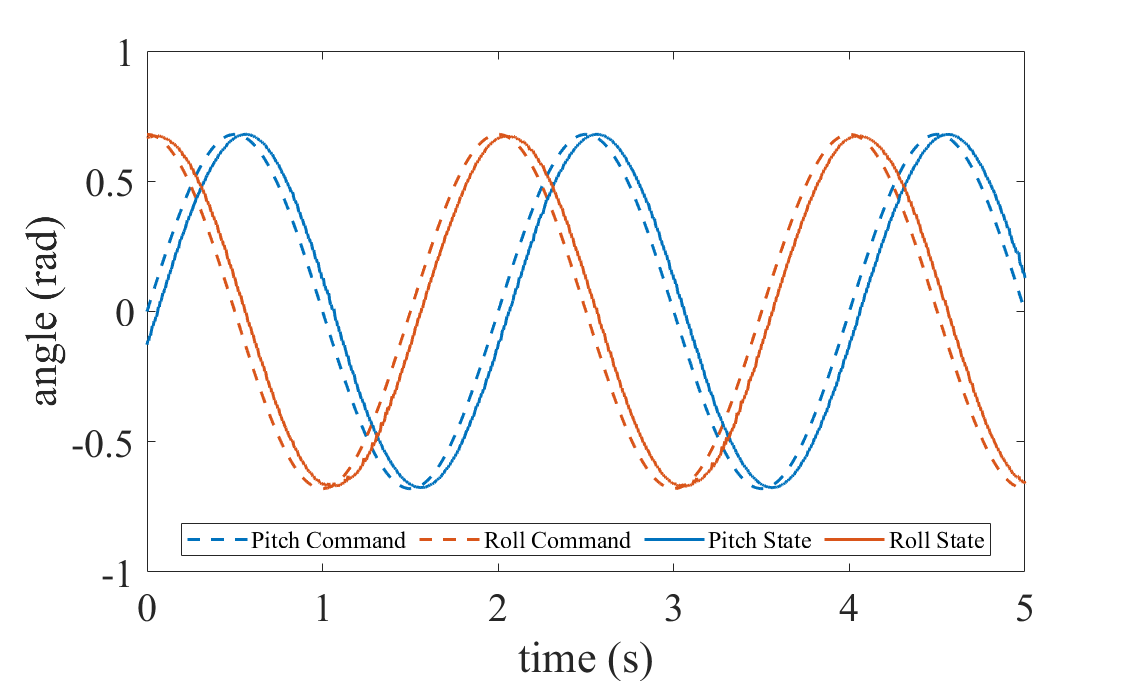}
        \caption{T = 2 s}
        \label{fig:4-10}
    \end{subfigure}
    \hfill
    \begin{subfigure}[t]{0.32\textwidth}
        \centering
        \includegraphics[width=\textwidth]{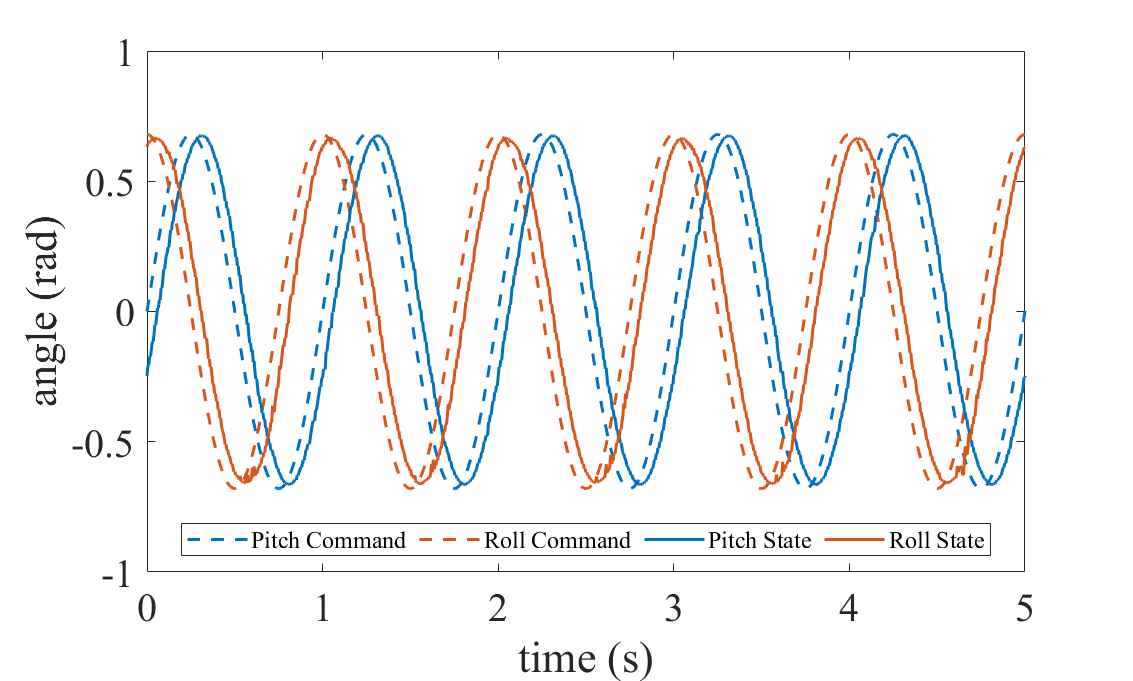}
        \caption{T = 1 s}
        \label{fig:4-11}
    \end{subfigure}

    \vspace{4pt}
    
    \begin{subfigure}[t]{0.24\textwidth}  
        \centering
        \includegraphics[width=\textwidth]{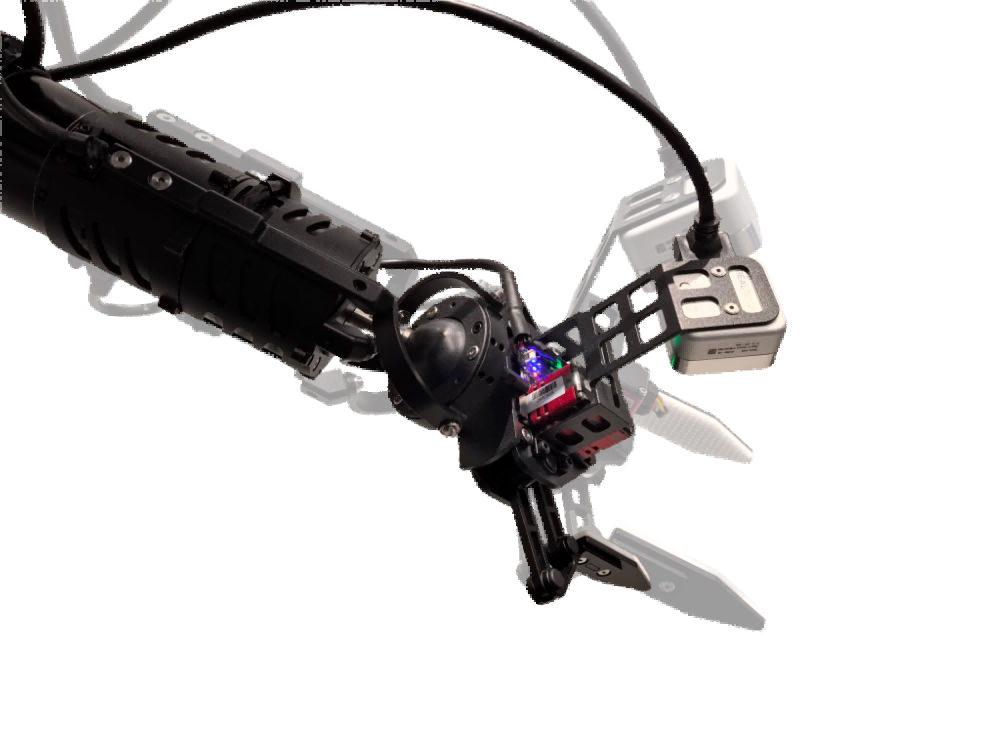}
        \caption{Pitch=0.48, Roll=0.48}
        \label{fig:4-1}
    \end{subfigure}
    \hfill
    \begin{subfigure}[t]{0.24\textwidth}
        \centering
        \includegraphics[width=\textwidth]{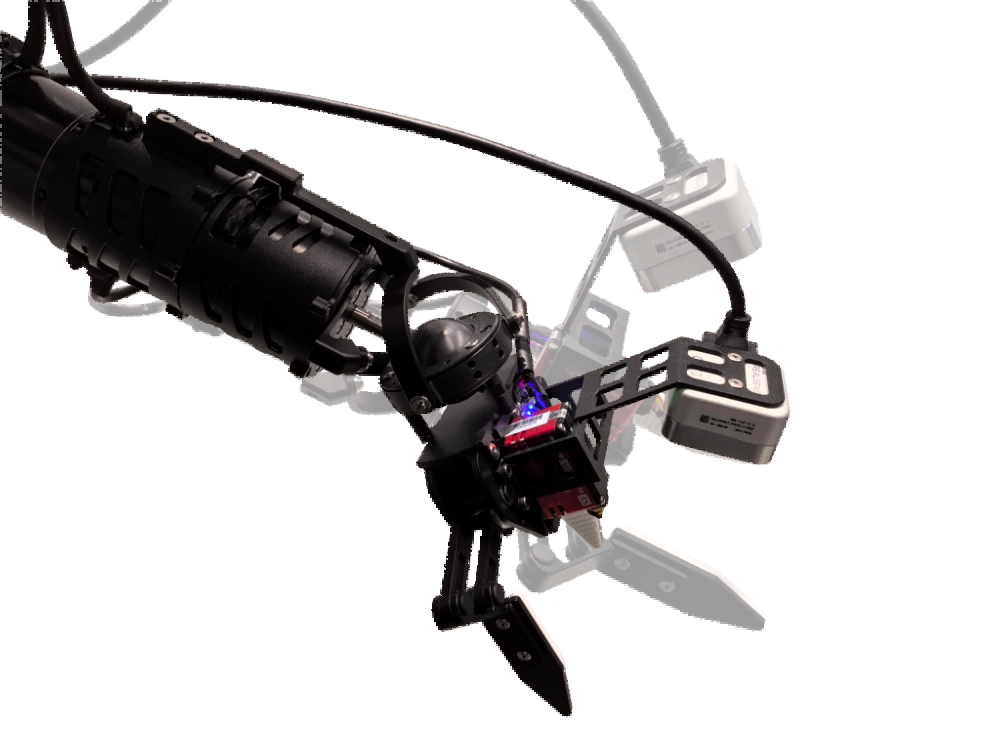}
        \caption{Pitch=0.68, Roll=0}
        \label{fig:4-2}
    \end{subfigure}
    \hfill
    \begin{subfigure}[t]{0.24\textwidth}
        \centering
        \includegraphics[width=\textwidth]{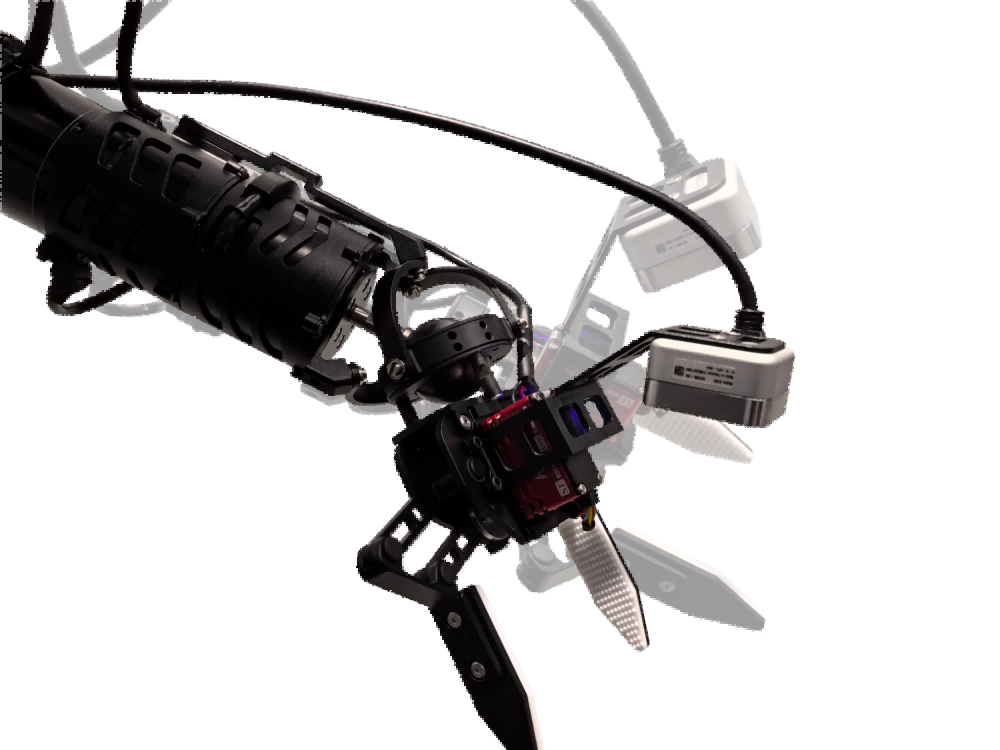}
        \caption{Pitch=0.48, Roll=-0.48}
        \label{fig:4-3}
    \end{subfigure}
    \hfill
    \begin{subfigure}[t]{0.24\textwidth}
        \centering
        \includegraphics[width=\textwidth]{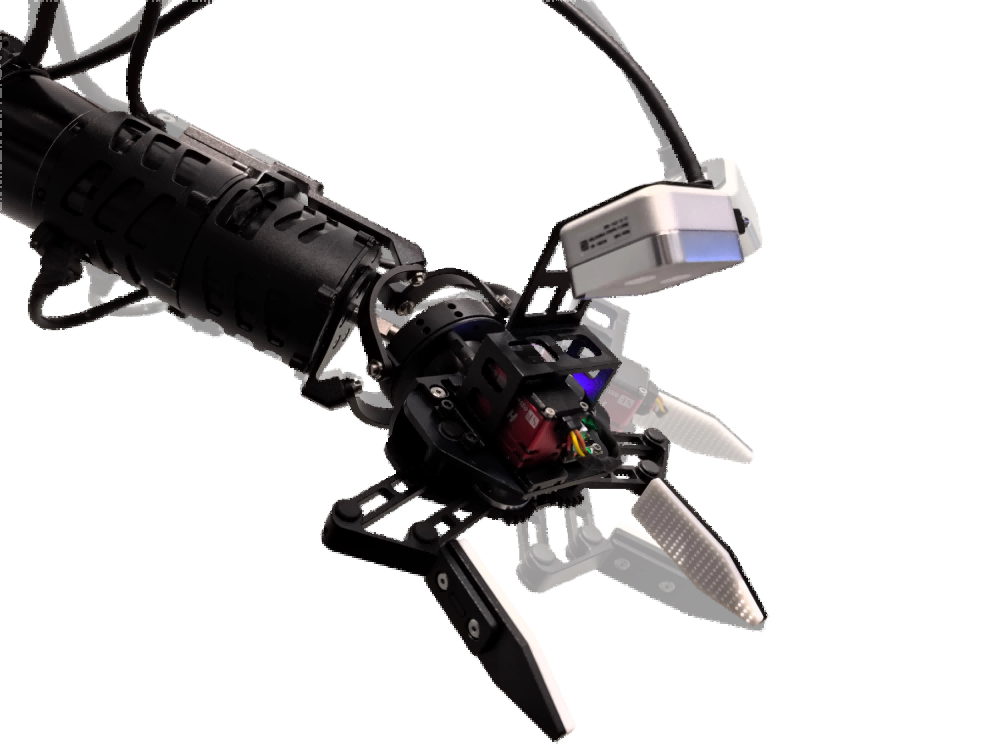}
        \caption{Pitch=0, Roll=-0.68}
        \label{fig:4-4}
    \end{subfigure}
    
    \vspace{4pt}
    
    \begin{subfigure}[t]{0.24\textwidth}
        \centering
        \includegraphics[width=\textwidth]{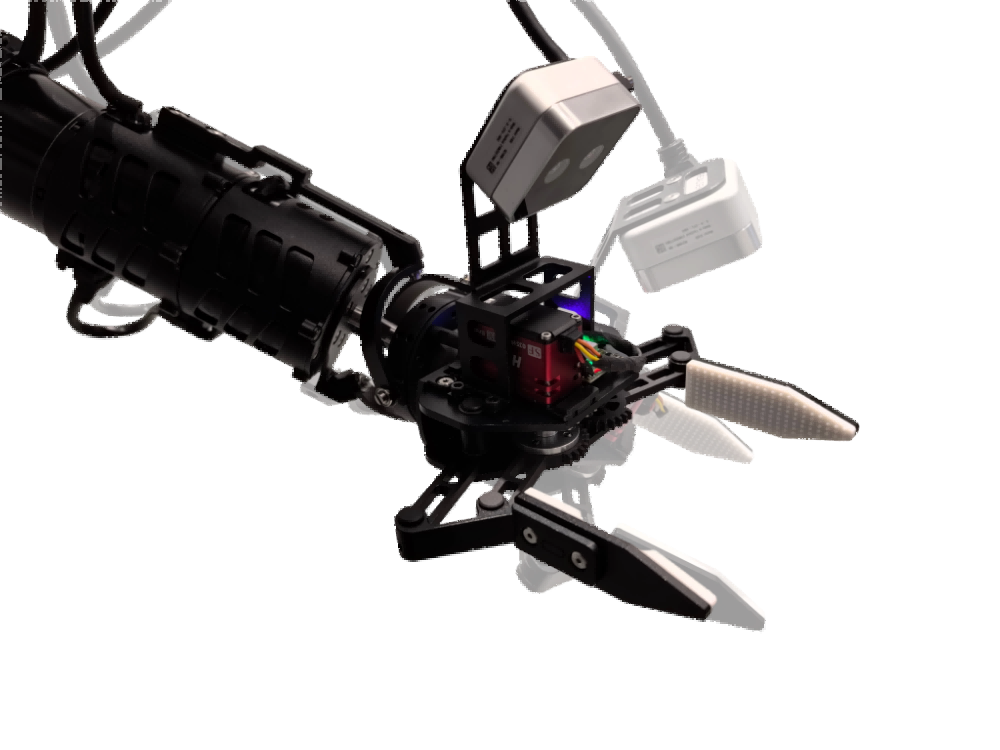}
        \caption{Pitch=-0.48, Roll=-0.48}
        \label{fig:4-5}
    \end{subfigure}
    \hfill
    \begin{subfigure}[t]{0.24\textwidth}
        \centering
        \includegraphics[width=\textwidth]{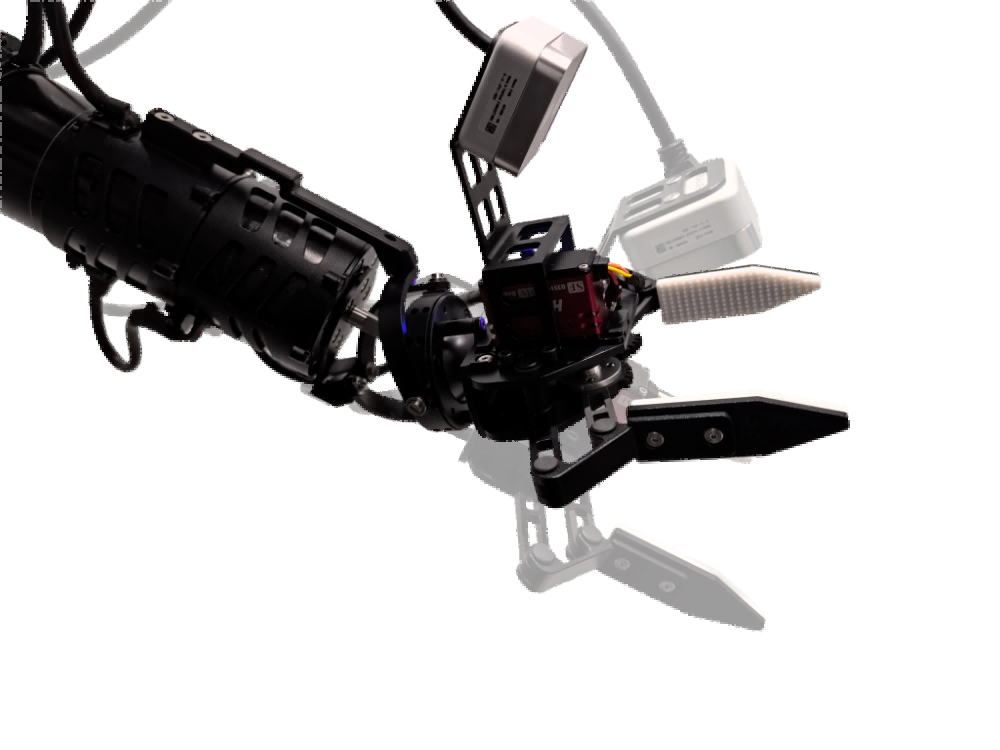}
        \caption{Pitch=-0.68, Roll=0}
        \label{fig:4-6}
    \end{subfigure}
    \hfill
    \begin{subfigure}[t]{0.24\textwidth}
        \centering
        \includegraphics[width=\textwidth]{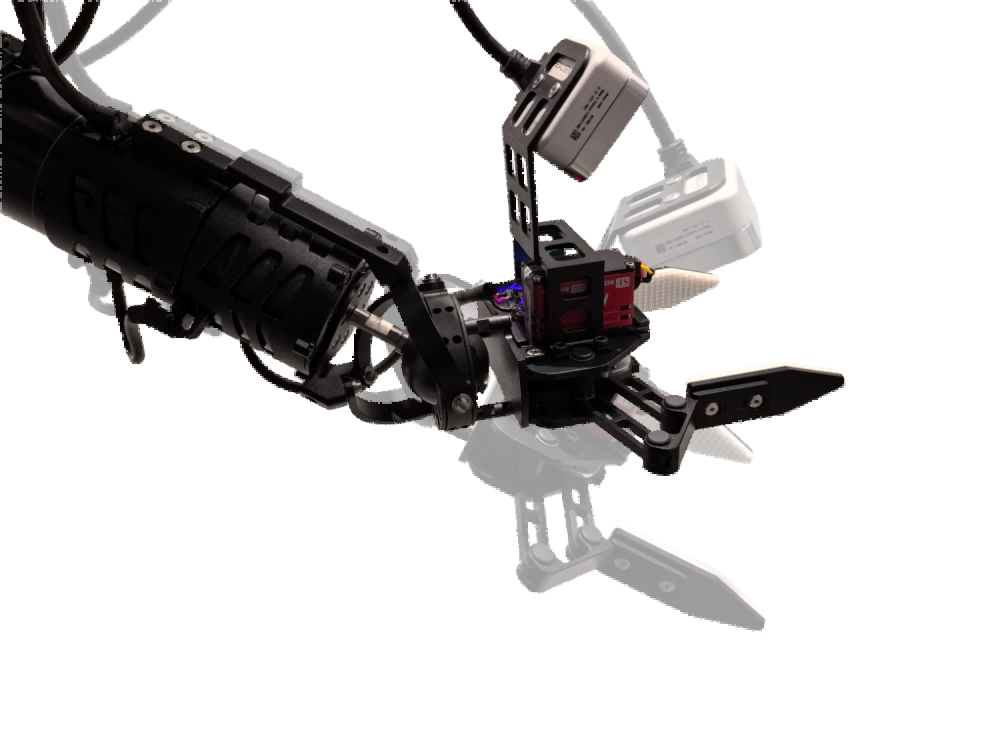}
        \caption{Pitch=-0.48, Roll=0.48}
        \label{fig:4-7}
    \end{subfigure}
    \hfill
    \begin{subfigure}[t]{0.24\textwidth}
        \centering
        \includegraphics[width=\textwidth]{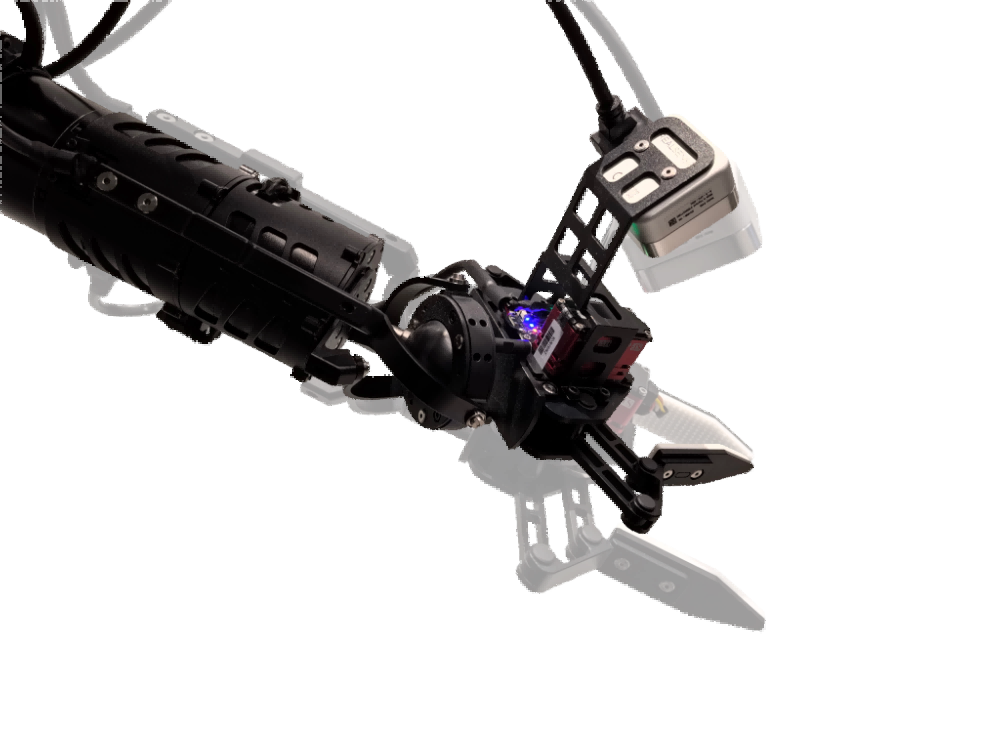}
        \caption{Pitch = 0, Roll = 0.68}
        \label{fig:4-8}
    \end{subfigure}

    \caption{Motion Range of ByteWrist, here we define moving down and left (along the direction of the gripper) as pitch and roll positive direction (Unit: Radian).}
    \label{fig:4}
\end{figure*}

As illustrated in Fig. \ref{fig:4}(a-c), the input of wrist movement and the actual position feedback are plotted for cycles \(T=4s\), \(2s\) and \(1s\) respectively. Postures of ByteWrist at 8 distinct moments are presented in Fig. \ref{fig:4}(d-k), each corresponding to different pitch and yaw angles. Throughout the wrist movement, the yaw direction remains constant. Experimental results demonstrate that the robotic wrist exhibits flexible motion within this operational region.

In the actual control of the wrist, the three aforementioned motion exhibit a delay of approximately 0.06\,s, attributed to communication and computation latencies. Owing to variations in motion speed, the errors between the state and command curves are 0.064 rad, 0.127 rad, and 0.247 rad, respectively, when the pitch or roll angle is 0 rad. Consequently, in high-dynamic applications, the influence induced by such lag must be considered, whereas it can be neglected in low-speed control scenarios.

\subsection{Confined-Space Maneuverability}
To verify the flexibility of ByteWrist in confined spaces, this study designs a comparative experiment of objects grasping inside a glove box between ByteMini and Kinova\cite{Kinova}.
 
\begin{figure}[h!]
    \centering
    \includegraphics[width=0.6\linewidth]{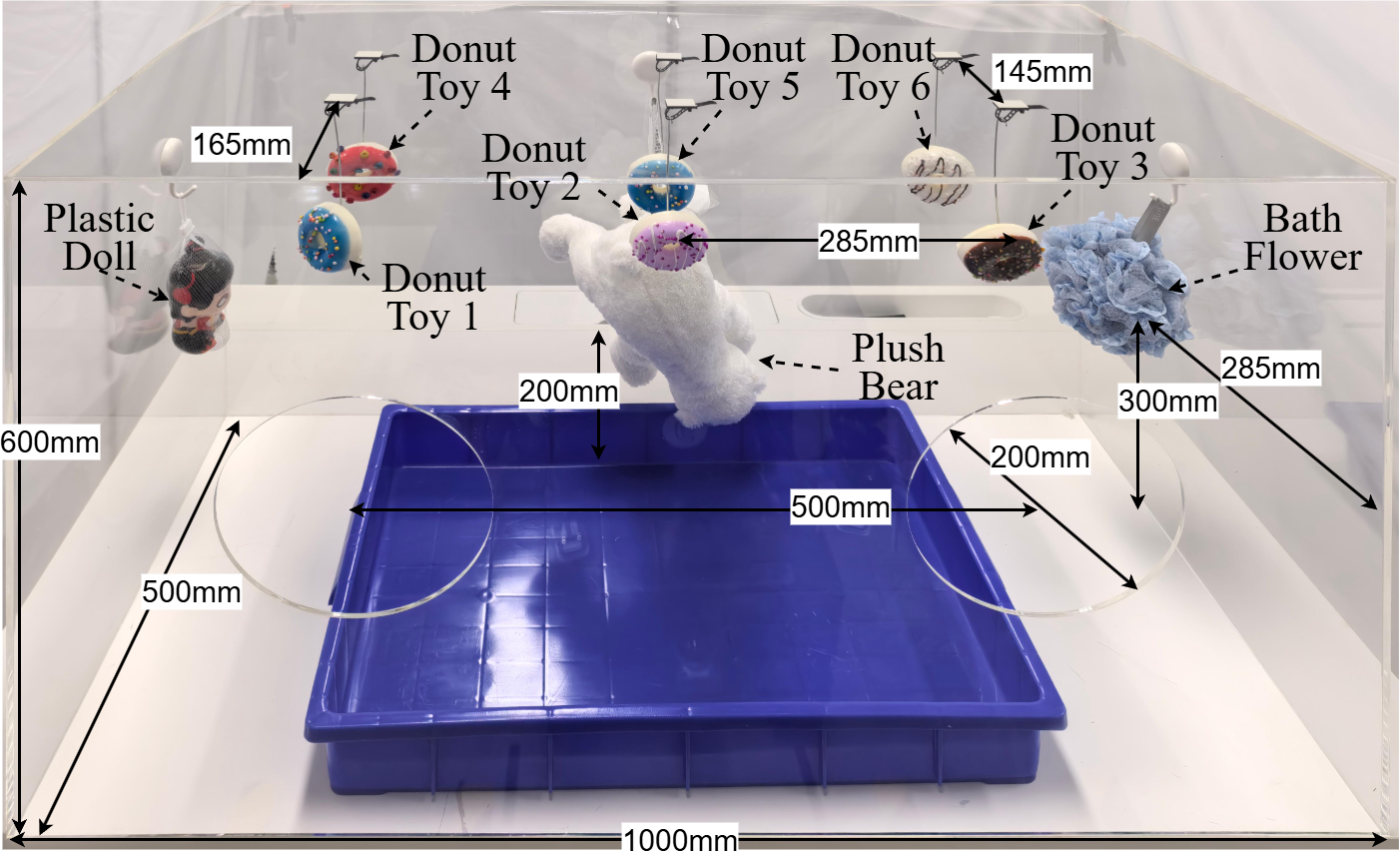}
    \caption{Glove Box Experimental Scenario Setup.}
    \label{fig:5}
\end{figure}

A glove box shown in Fig. \ref{fig:5} is designed for this experiment. This box is constructed from transparent acrylic material, with dimensions of 1000\,mm (length) × 500\,mm (width) × 600\,mm (height). There are two access holes (each in diameter 200\,mm) on the front panel of the cabinet, with center-to-center distance in 500\,mm. 

Nine grasping target objects were arranged inside the glove box, including a plastic doll, a plush bear, a bath flower, and six donut toys. The specific distribution of these objects in the glove box is illustrated in Fig. \ref{fig:5}. The design objective is to evaluate the difficulty of grasping tasks in different spatial regions through two access holes, thereby achieving the assessment of the maneuverability of robotic arms.

\begin{figure*}[thpb]
    \vspace{5pt}
    \centering
    \setlength{\tabcolsep}{4pt} 
    
    \begin{subfigure}[t]{0.32\textwidth}  
        \centering
        \includegraphics[width=\textwidth]{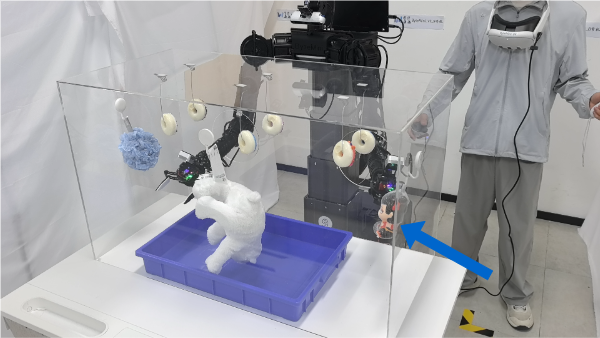}
        \caption{Plastic Doll}
        \label{fig:6-1}
    \end{subfigure}
    \hfill
    \begin{subfigure}[t]{0.32\textwidth}
        \centering
        \includegraphics[width=\textwidth]{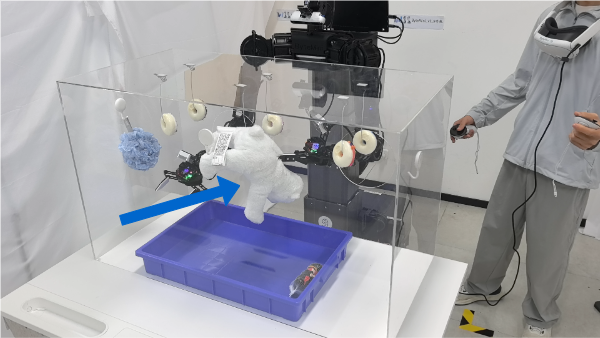}
        \caption{Plush Bear}
        \label{fig:6-2}
    \end{subfigure}
    \hfill
    \begin{subfigure}[t]{0.32\textwidth}
        \centering
        \includegraphics[width=\textwidth]{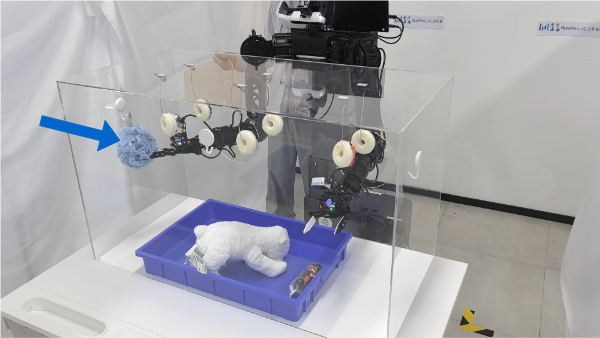}
        \caption{Bath Flower}
        \label{fig:6-3}
    \end{subfigure}
    
    \vspace{4pt}
    
    \begin{subfigure}[t]{0.32\textwidth}
        \centering
        \includegraphics[width=\textwidth]{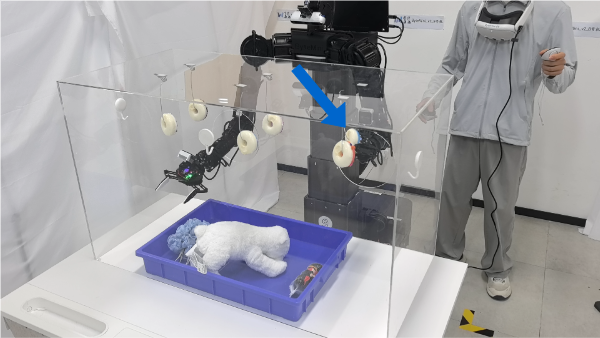}
        \caption{Donut Toy 1}
        \label{fig:6-4}
    \end{subfigure}
    \hfill
    \begin{subfigure}[t]{0.32\textwidth}
        \centering
        \includegraphics[width=\textwidth]{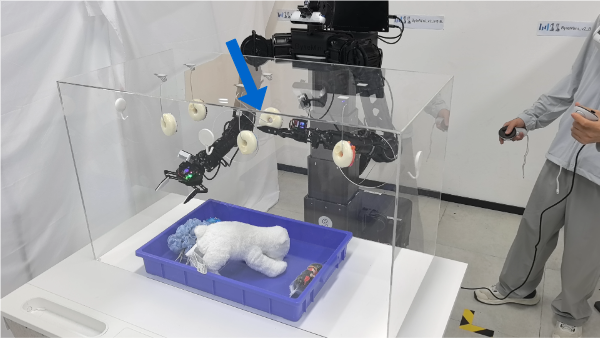}
        \caption{Donut Toy 2}
        \label{fig:6-5}
    \end{subfigure}
    \hfill
    \begin{subfigure}[t]{0.32\textwidth}
        \centering
        \includegraphics[width=\textwidth]{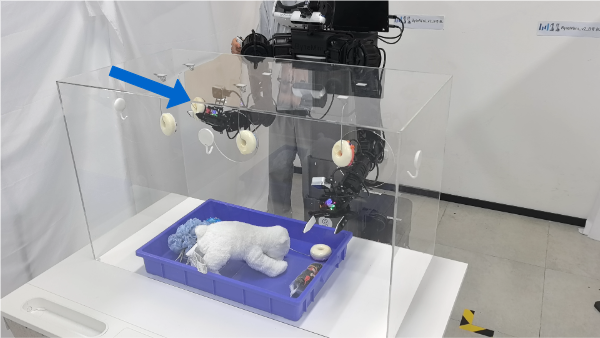}
        \caption{Donut Toy 3}
        \label{fig:6-6}
    \end{subfigure}
    
    \vspace{4pt}
    
    \begin{subfigure}[t]{0.32\textwidth}  
        \centering
        \includegraphics[width=\textwidth]{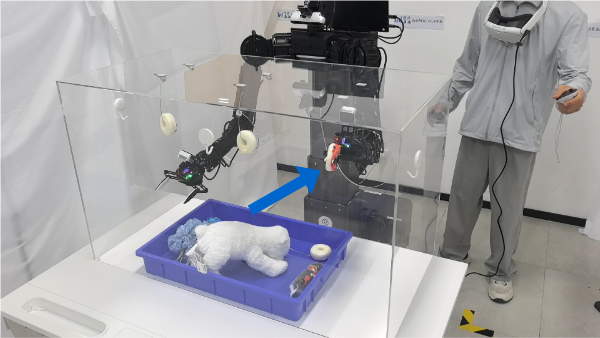}
        \caption{Donut Toy 4}
        \label{fig:6-7}
    \end{subfigure}
    \hfill
    \begin{subfigure}[t]{0.32\textwidth}
        \centering
        \includegraphics[width=\textwidth]{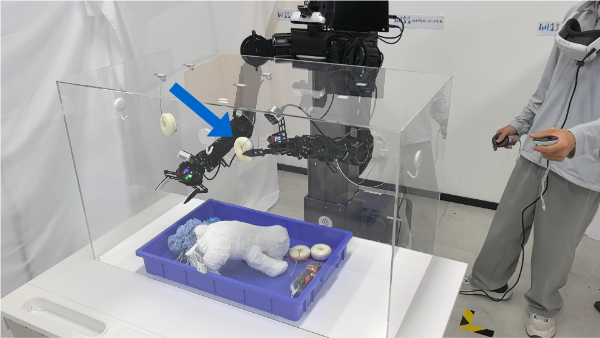}
        \caption{Donut Toy 5}
        \label{fig:6-8}
    \end{subfigure}
    \hfill
    \begin{subfigure}[t]{0.32\textwidth}
        \centering
        \includegraphics[width=\textwidth]{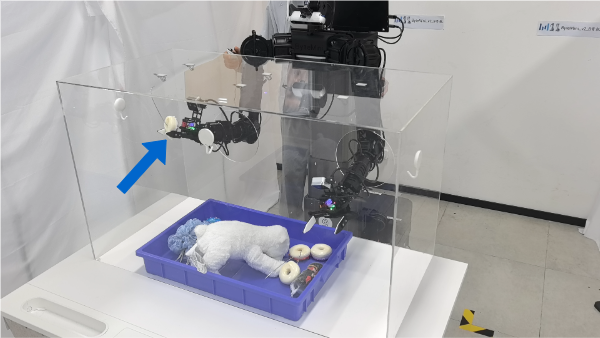}
        \caption{Donut Toy 6}
        \label{fig:6-9}
    \end{subfigure}
    
    \caption{ByteMini Performance in the Confined-Space Grasping Experiment.}
    \label{fig:6}
\end{figure*}

The grasping experiment procedure is as follows. The ByteMini robot is tele-operated to extend both arms into the glove box. Subsequently, its waist and mobile chassis are kept stationary, and the current pose of the arms are recorded as the initial pose. Next, the arms are tele-operated via Quest VR to attempt grasping the objects inside the glove box and placing them into the blue storage basket. Once the object is placed into the blue storage box, the robot arm is subsequently returned to its initial pose in preparation for the next grasping attempt. Throughout the procedure, the duration to grasp each object (between two initial poses) is recorded. The grasping sequence is as follows: plastic doll, plush bear, bath flower, and Donut Toys 1–6. As shown in Fig. \ref{fig:6}, the ByteMini robot with ByteWrist is capable of achieving accurate grasping of all nine objects, with the time required for each grasp summarized in Table \ref{table_2}. 

\begin{table}[h!]
    \centering
    \begin{subtable}[t]{0.45\linewidth}  
        \centering
        \caption{Grasping Time by ByteMini (Unit:\,s)}
        \label{table_2}
        \begin{tabular}{ccc}
            \toprule
            Plastic Doll & Plush Bear & Bath Flower \\
            \midrule
            20 & 49 & 21 \\
            \midrule
            Donut Toy 1 & Donut Toy 2 & Donut Toy 3 \\
            \midrule
            22 & 38 & 23 \\
            \midrule
            Donut Toy 4 & Donut Toy 5 & Donut Toy 6 \\
            \midrule
            17 & 27 & 17 \\
            \midrule
            \multicolumn{3}{c}{Total Time} \\
            \midrule
            \multicolumn{3}{c}{234} \\
            \bottomrule
        \end{tabular}
    \end{subtable}
    \hfill  
    \begin{subtable}[t]{0.45\linewidth}  
        \centering
        \caption{Grasping Time by Kinova (Unit:\,s)}
        \label{table_3}
        \begin{tabular}{ccc}
            \toprule
            Plastic Doll & Plush Bear & Bath Flower \\
            \midrule
            47 & 56 & 67 \\
            \midrule
            Donut Toy 1 & Donut Toy 2 & Donut Toy 3 \\
            \midrule
            64 & 51 & 75 \\
            \midrule
            Donut Toy 4 & Donut Toy 5 & Donut Toy 6 \\
            \midrule
            23 & 59 & 34 \\
            \midrule
            \multicolumn{3}{c}{Total Time} \\
            \midrule
            \multicolumn{3}{c}{476} \\
            \bottomrule
        \end{tabular}
    \end{subtable}
    \caption{Grasping Time Comparison Between ByteMini and Kinova}  
    \label{table:combined}  
\end{table}

\begin{figure}[htb!]
    \centering
    \includegraphics[width=0.5\linewidth]{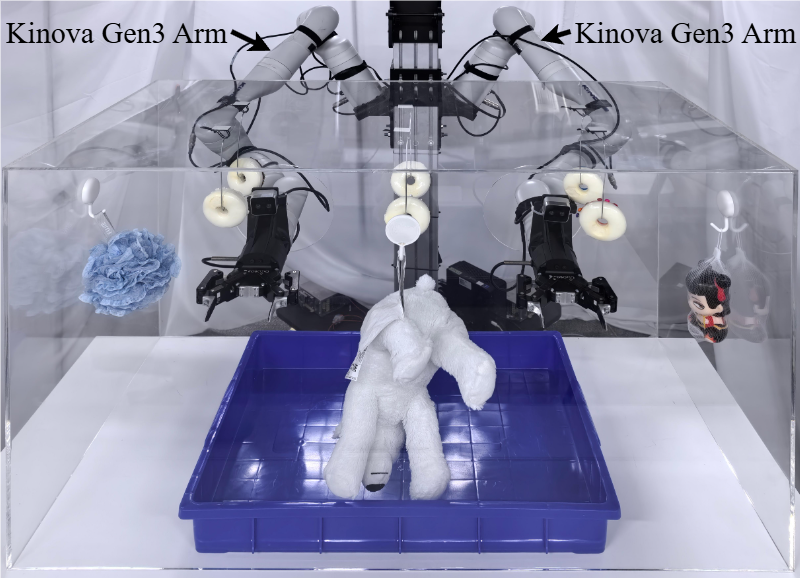}
    \caption{The Kinova Dual-arm Robot Setup.}
    \label{fig:7}
\end{figure}

To enable comparison with the traditional serial wrist, a dual-arm robot featuring a fixed base is constructed in this study, as illustrated in Fig. \ref{fig:7}. The robot two arms consist of Kinova Gen3, which utilizes a 7-DoF SRS configuration similar to ByteMini. The key distinction lies in Kinova adopting a serial configuration for the wrist three joints. A Robotiq 2F-85 gripper is affixed as the end effector of each arm. The identical grasping experiment is conducted using the Kinova dual-arm robot, with grasping times recorded, and the experimental results are summarized in Table \ref{table_3}.

Between Table \ref{table_2} and \ref{table_3}, we can notice that for the same experiment task, the Kinova dual-arm robot takes approximately twice as long as ByteMini, which reflects the difference in the flexibility of the robots wrist. Three main reasons lead to Kinova's longer grasping duration are as follows.

\begin{figure*}[tb!]
    \vspace{5pt}
    \centering
    \setlength{\tabcolsep}{4pt} 
    
    \begin{subfigure}[t]{0.32\textwidth} 
        \centering
        \includegraphics[width=\textwidth, trim=0 9pt 0 0, clip]{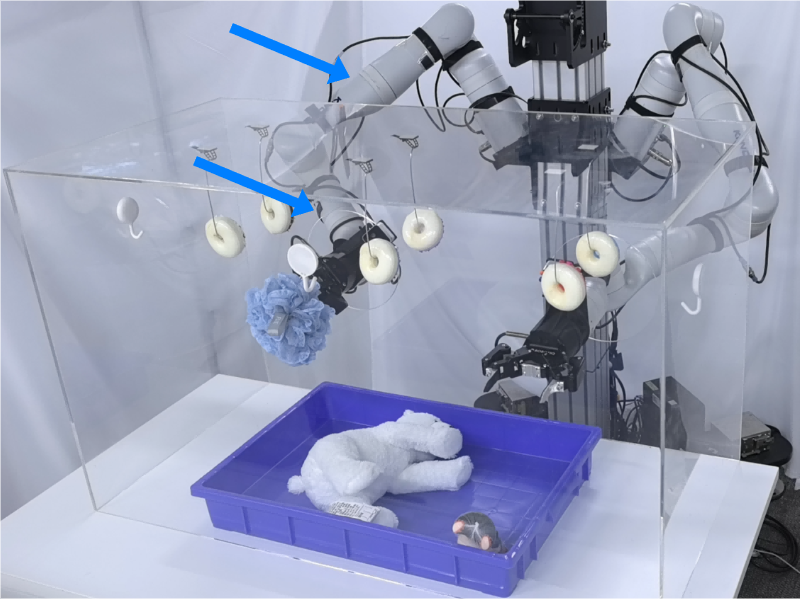}
        \caption{Larger Arm Angle Adjustment.}
        \label{fig:8-1}
    \end{subfigure}
    \hfill
    \begin{subfigure}[t]{0.32\textwidth}
        \centering
        \includegraphics[width=\textwidth, trim=0 9pt 0 0, clip]{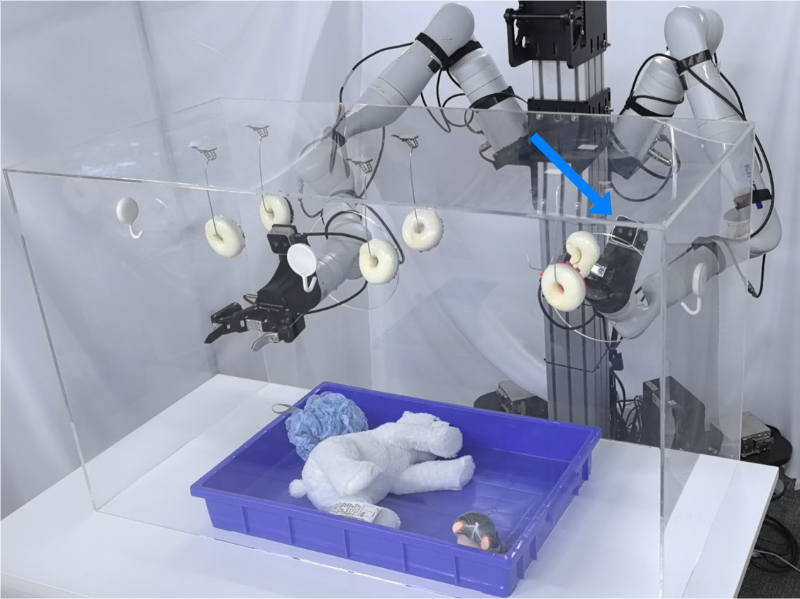}
        \caption{Gripper Camera Out of the Box.}
        \label{fig:8-2}
    \end{subfigure}
    \hfill
    \begin{subfigure}[t]{0.32\textwidth}
        \centering
        \includegraphics[width=\textwidth]{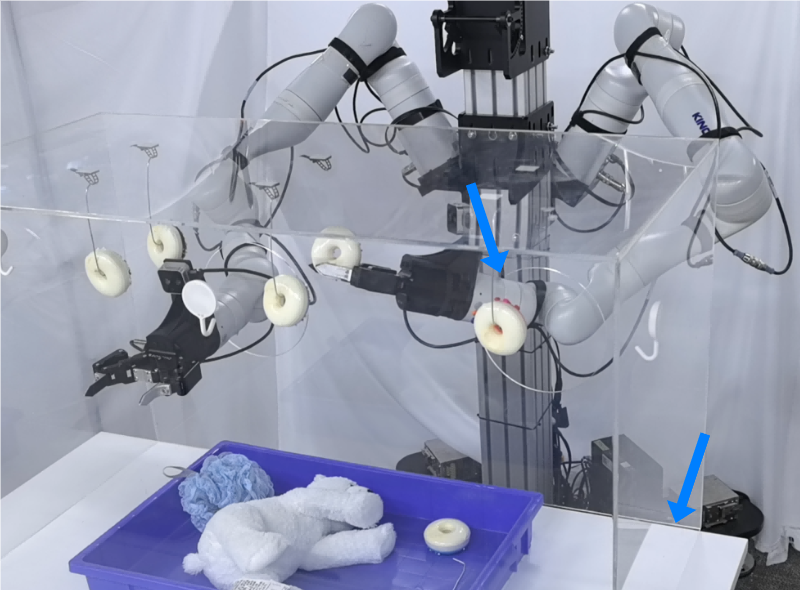}
        \caption{Slight Collisions Occur.}
        \label{fig:8-3}
    \end{subfigure}

    \caption{Causes of Longer Grasping Duration for Kinova Dual-Arm Robot in the Confined Space Grasping Experiment.}
    \label{fig:8}
\end{figure*}

\begin{enumerate}
    \item \textbf{Larger Arm Angle Adjustment:} For the grasping of the plastic doll, bath flower, donut toy 4, and donut toy 6, the Kinova's forearm—featuring a serial configuration and greater length—requires retraction from the glove box to adjust its posture prior to grasp completion. As illustrated in Fig. \ref{fig:8-1}, to compensate for the limited range of wrist rotation, the upper arm executes a large-angle rotational motion, resulting in increased grasping duration.
    \item \textbf{Gripper Camera Out of the Box:} When attempting to grasp donut toy 1 and donut toy 3, the rotation of the Kinova's wrist will cause the collision between the gripper camera and the inner front wall of the glove box, if the gripper camera remains in the glove box. Thus, the gripper camera must first be moved out of the glove box before grasping, which leads to longer grasping time.
    \item \textbf{Slight Collisions Occur:} For the grasping of donut toy 2 and donut toy 5,  slight collisions occur between the forearm links and the glove box due to the forearm's series configuration and greater length, which brings more time for adjustment and recovery.
\end{enumerate}

\subsection{Dual-Arm Manipulation of Deformable Objects}
GR-3\cite{GR3, GR2} is a large-scale vision-language-action (VLA) model\cite{vla, vla1, vla2, vla3}, and ByteMini robots are deployed for GR-3 data collection and model rollout. The dexterous clothes-hanging task in GR-3 imposes high requirements on the capabilities of the robot, where the robot is required to achieve dual-arm collaboration in the chest area and perform manipulation of deformable objects with high precision and high dexterity.

\begin{figure*}[h!]
    \vspace{5pt}
    \centering
    \setlength{\tabcolsep}{4pt} 
    
    \begin{subfigure}[t]{0.32\textwidth}  
        \centering
        \includegraphics[width=\textwidth]{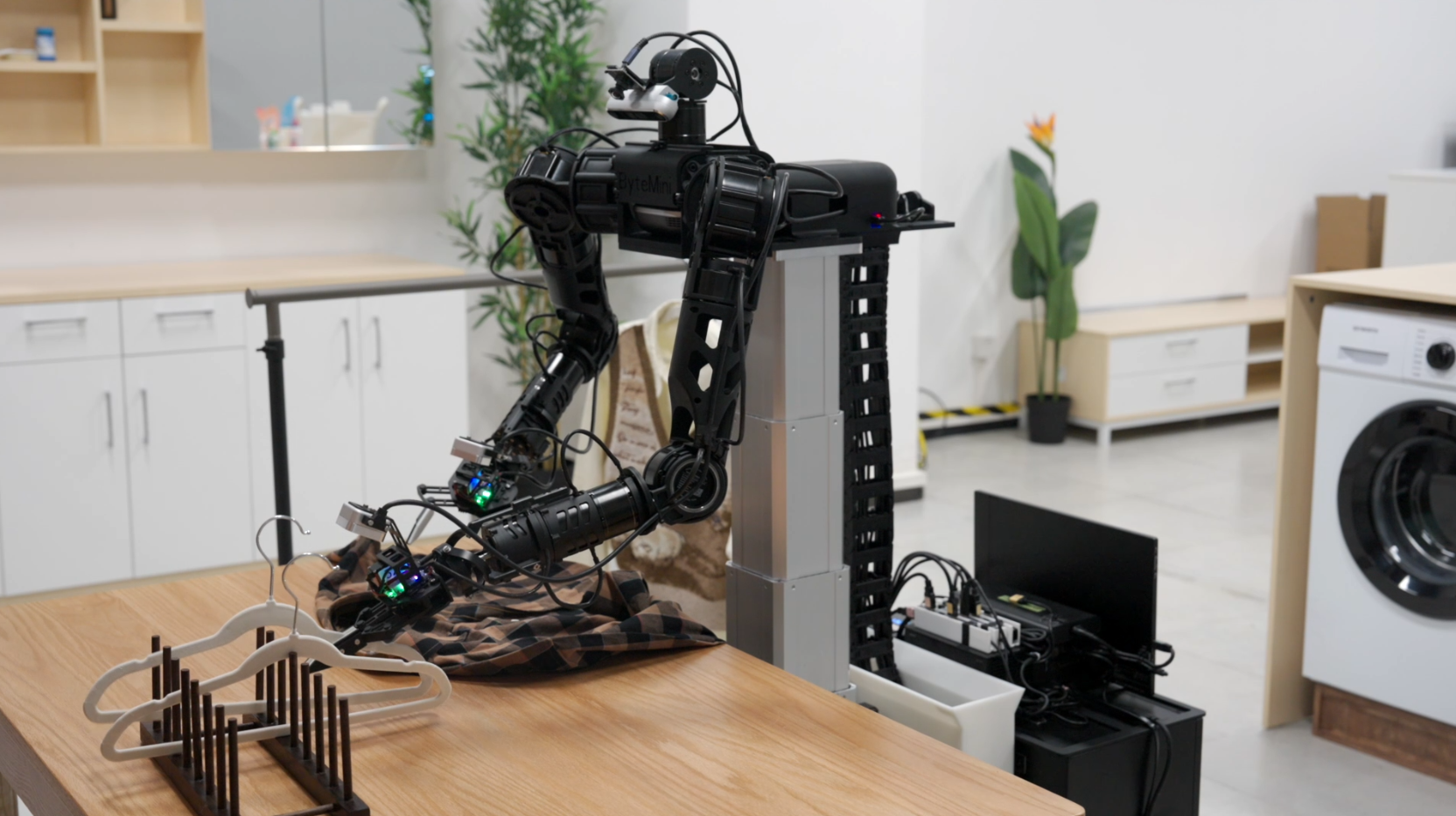}
        \caption{Pick Up the Clothes Hanger.}
        \label{fig:9-1}
    \end{subfigure}
    \hfill
    \begin{subfigure}[t]{0.32\textwidth}
        \centering
        \includegraphics[width=\textwidth]{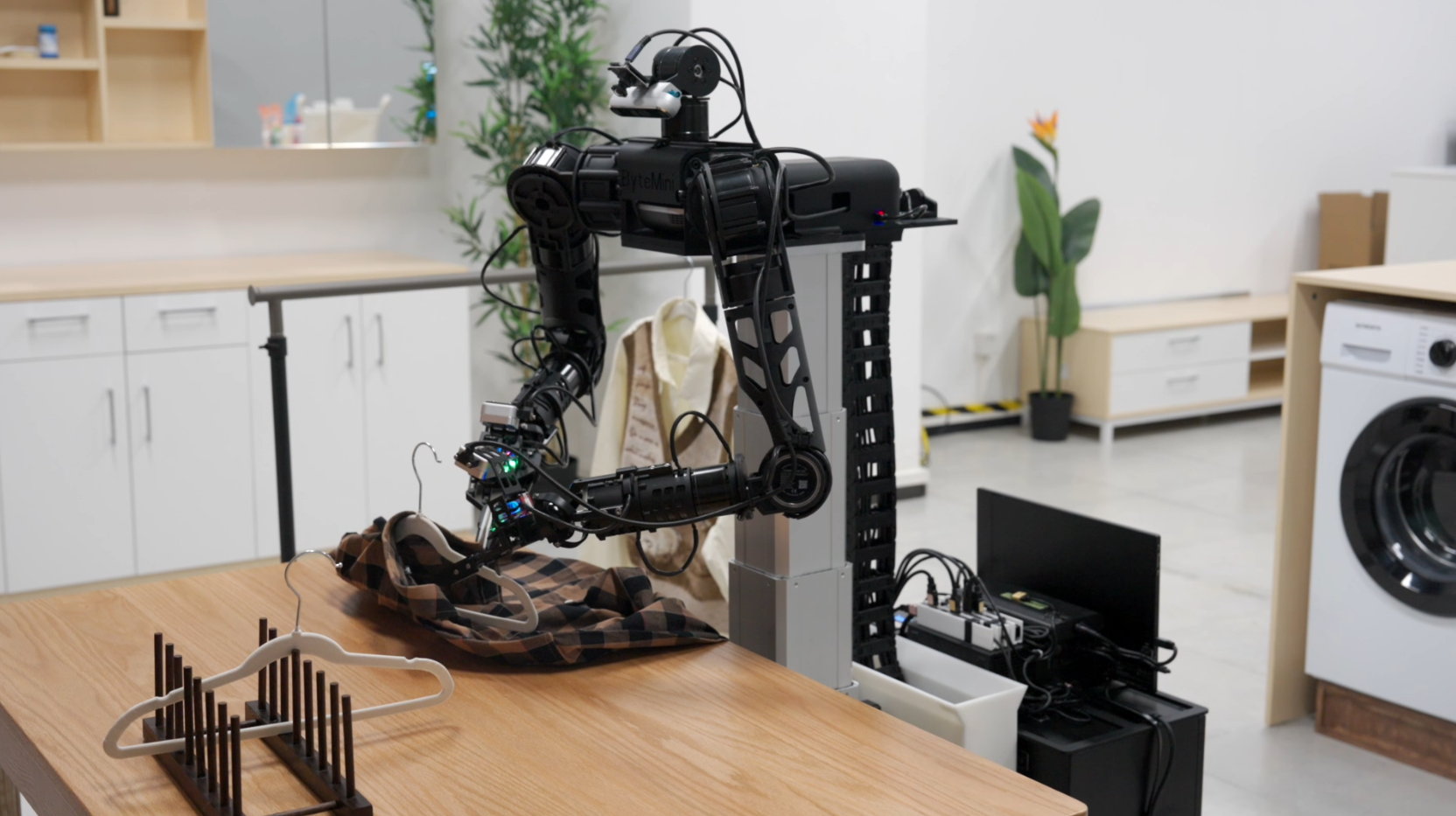}
        \caption{Hang the Clothes Left Shoulder.}
        \label{fig:9-2}
    \end{subfigure}
    \hfill
    \begin{subfigure}[t]{0.32\textwidth}
        \centering
        \includegraphics[width=\textwidth]{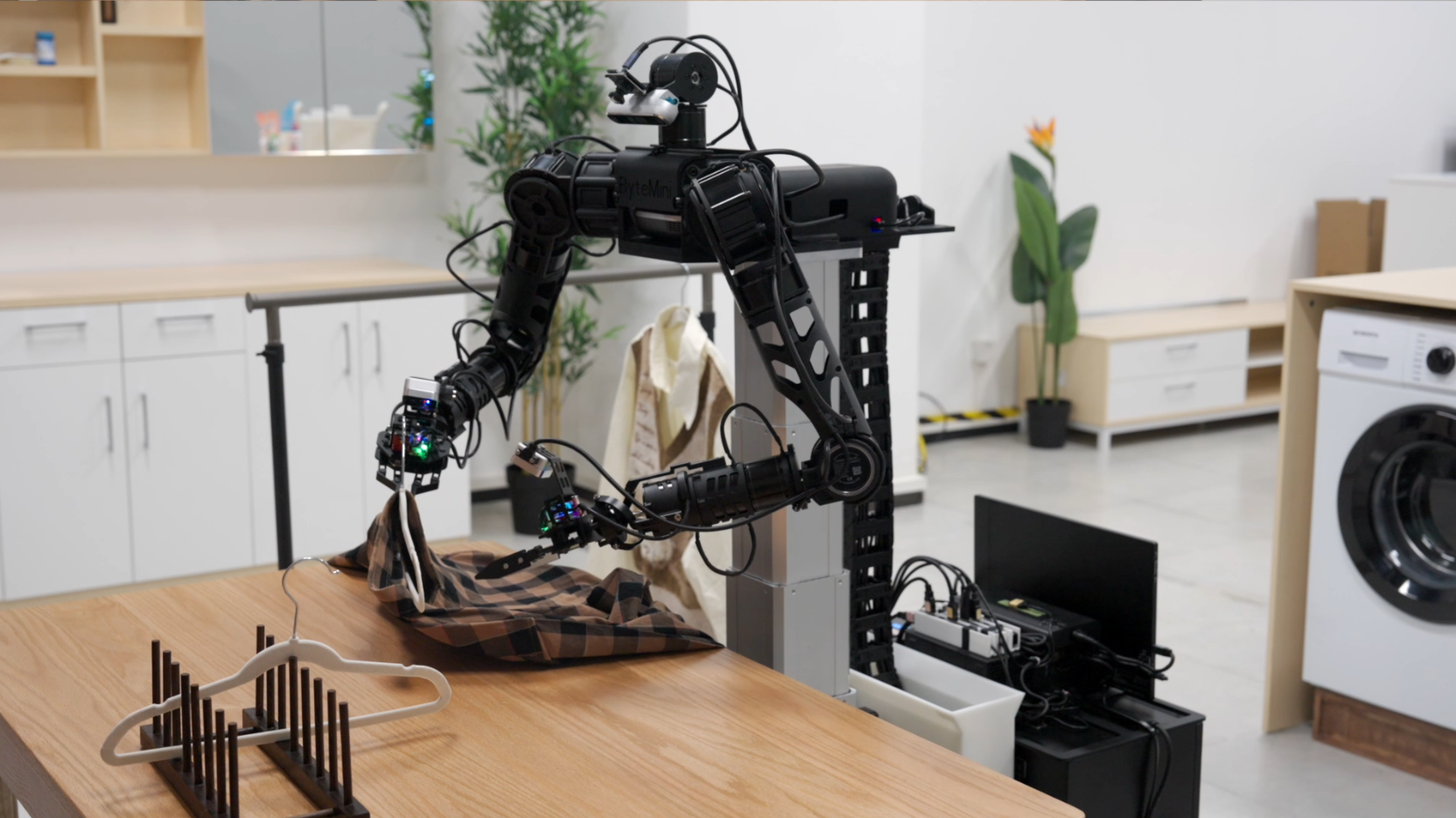}
        \caption{Switch the Holding Gripper.}
        \label{fig:9-3}
    \end{subfigure}
    
    \vspace{4pt}
    
    \begin{subfigure}[t]{0.32\textwidth}
        \centering
        \includegraphics[width=\textwidth]{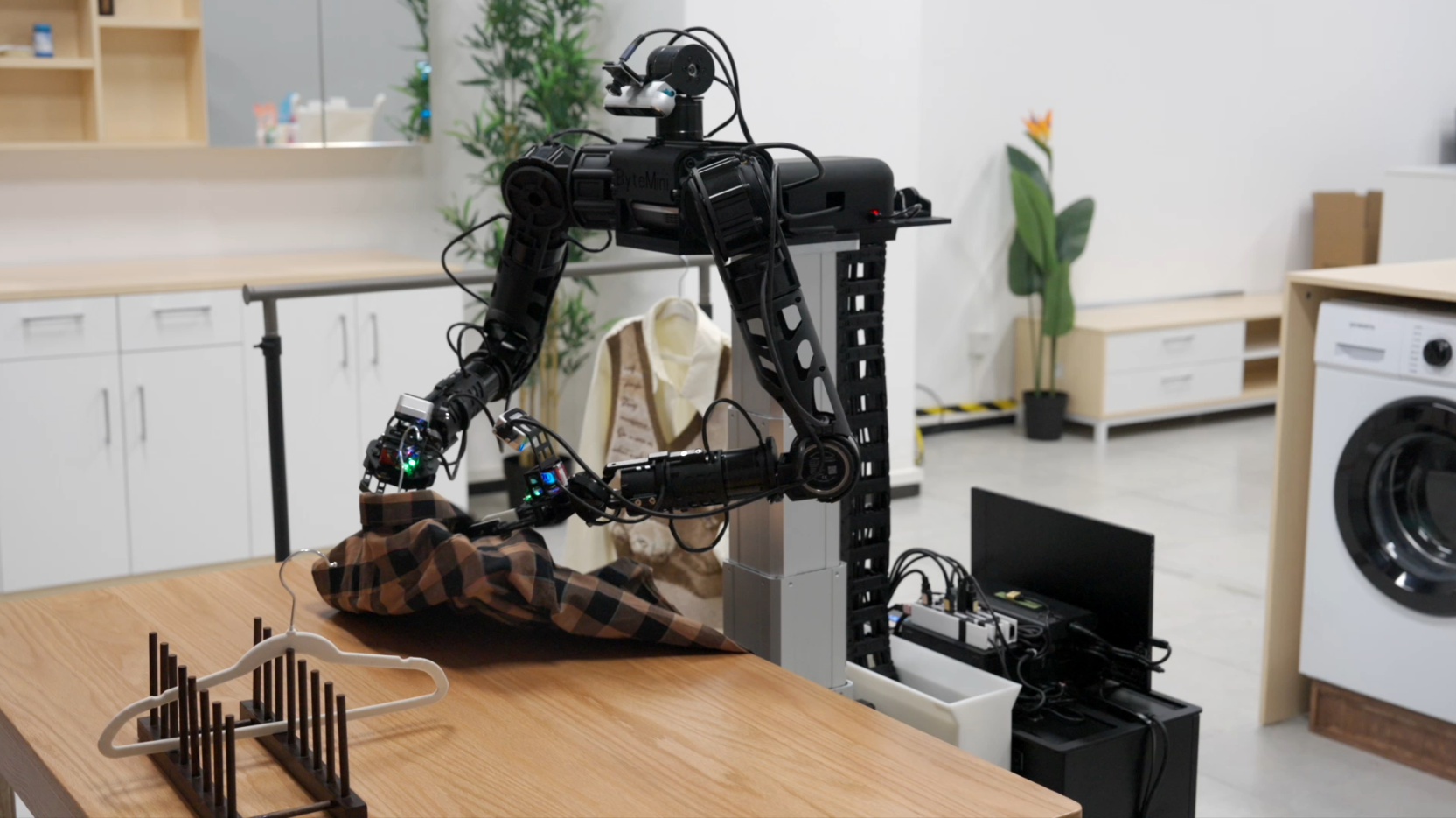}
        \caption{Hang the Clothes Right Shoulder.}
        \label{fig:9-4}
    \end{subfigure}
    \hfill
    \begin{subfigure}[t]{0.32\textwidth}
        \centering
        \includegraphics[width=\textwidth]{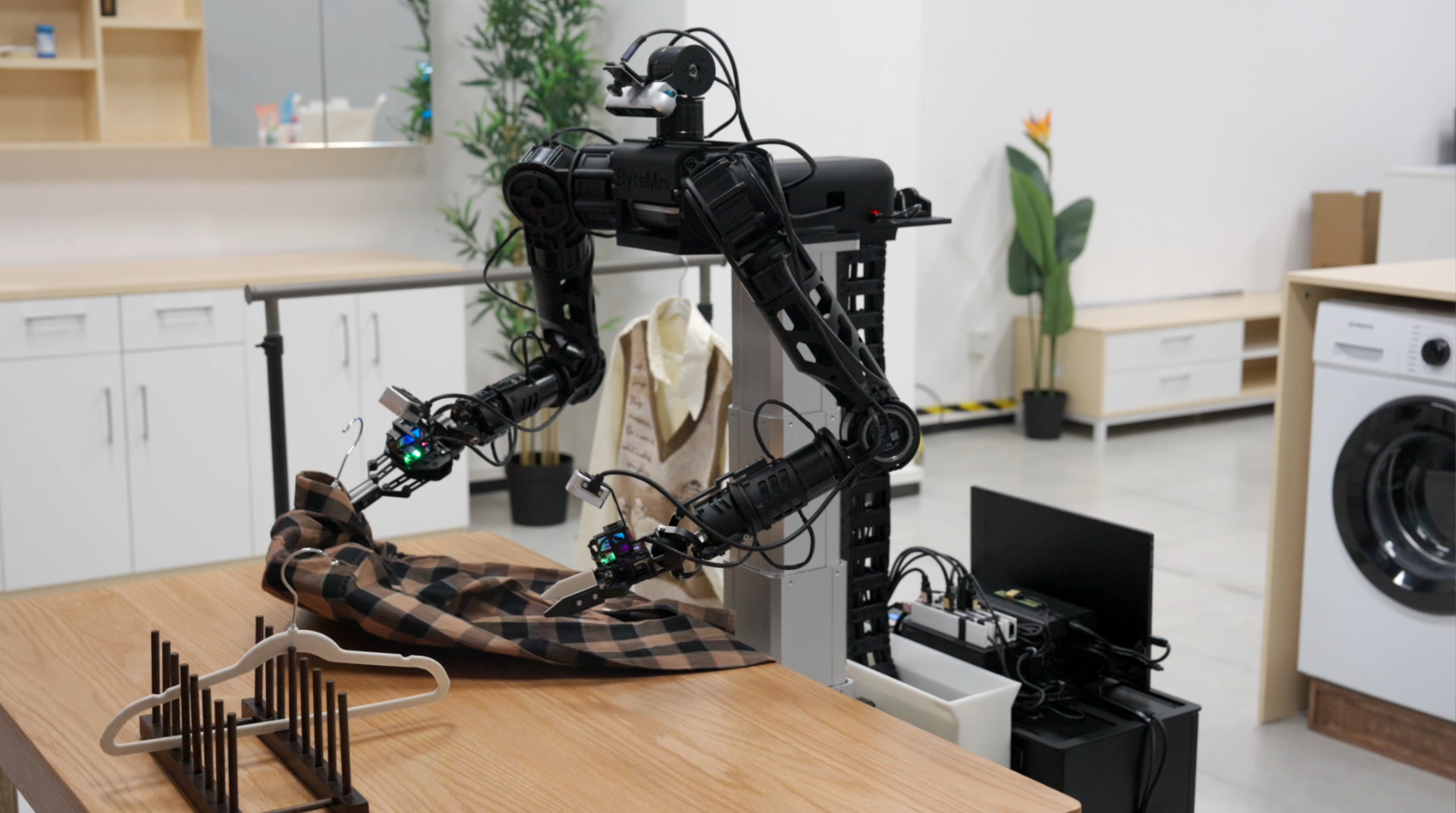}
        \caption{Pick Up the Assembly.}
        \label{fig:9-5}
    \end{subfigure}
    \hfill
    \begin{subfigure}[t]{0.32\textwidth}
        \centering
        \includegraphics[width=\textwidth]{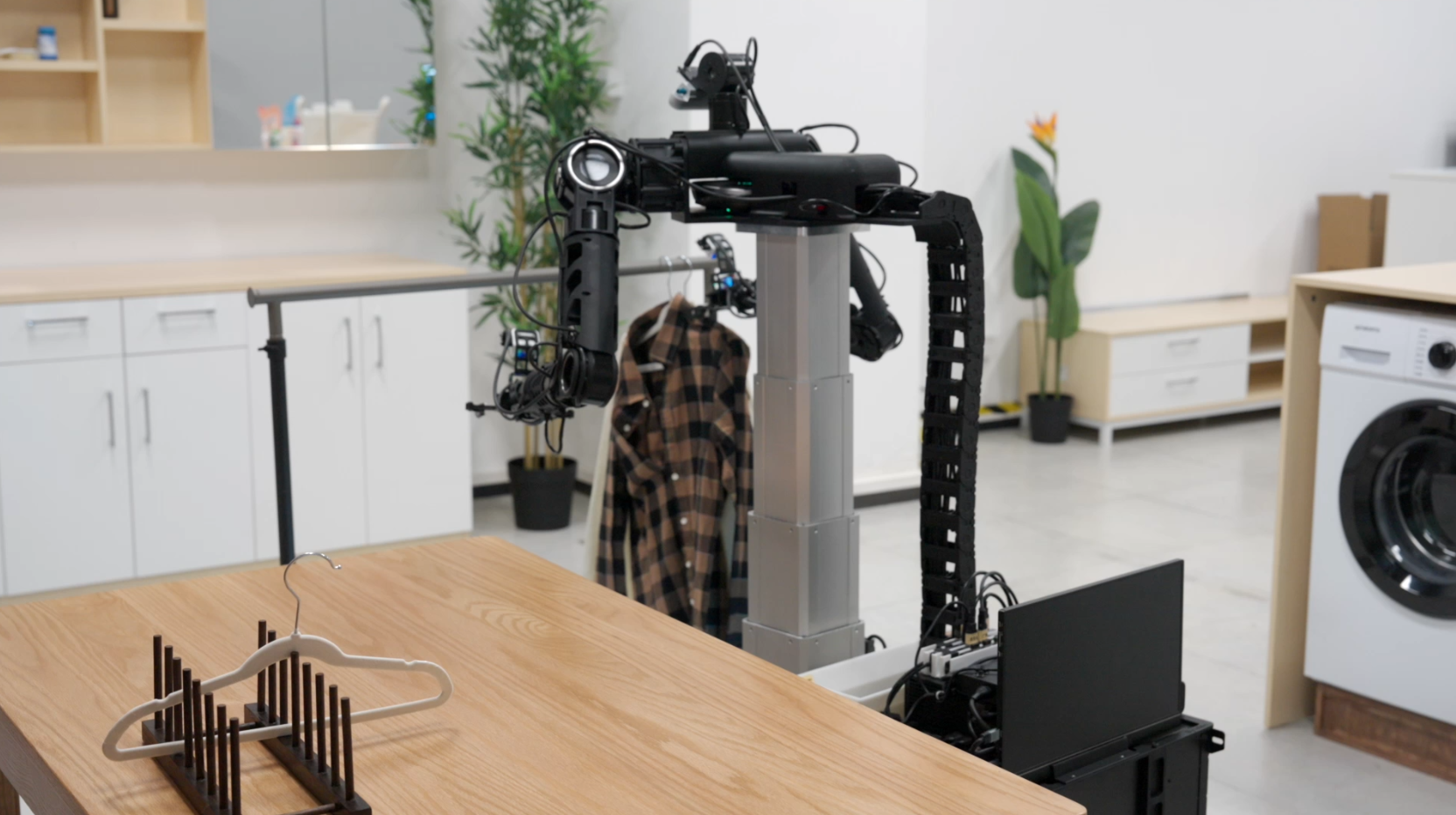}
        \caption{Hang the Assembly on the Rail.}
        \label{fig:9-6}
    \end{subfigure}

    \caption{GR-3 Task: ByteWrist Based Dual-Arm Manipulation of Deformable Objects.}
    \label{fig:9}
\end{figure*}

As shown in Fig. \ref{fig:9}, ByteMini is required to perform following actions in the clothes-hanging manipulation: pick up the clothes hanger with its left gripper, cooperate with both arms to hang the clothes left shoulder on the hanger, switch to holding the hanger with its right gripper, cooperate with both arms to hang the clothes right shoulder on the hanger, and hang the entire assembly onto the crossbar.

ByteMini has not only completed 116 hours of data collection for the clothes-hanging task, but also successfully achieved fully automated dexterous cloth manipulation, demonstrating the flexibility and robustness of ByteWrist based robotic arms. Last but not least, thanks to the design of ByteWrist, the clothes-hanging motion of the ByteMini robot exhibits remarkable antropomorphic, with the overall movement being smooth and natural, as illustrated in Fig. \ref{fig:9}.

\section{CONCLUSIONS}
This paper presents the design, modeling, and experimental validation of ByteWrist, a compact parallel robotic wrist developed for flexible operation in narrow and confined spaces. The research achievements and key conclusions are summarized as follows:

\begin{enumerate}
    \item \textbf{Innovative Structural Design:} ByteWrist adopts a three-stage motor-driven parallel mechanism, combined with arc-shaped end linkages and a central supporting ball. This design achieves a balance between compactness and stiffness. The prototype test confirms that the wrist can stably realize continuous RPY motion, meeting the requirements of anthropomorphic manipulation in confined spaces.
    \item \textbf{Complete Kinematic Modeling:} The forward and inverse kinematic models of ByteWrist are established. For forward kinematics, the Newton-Raphson method is used to solve the nonlinear equations, realizing the mapping from driving linkage angles \(\theta_1, \theta_2, \theta_3\) to parallel platform RPY angles. For inverse kinematics, the rotation matrix is derived based on given RPY angles to calculate the required driving linkage angles. Additionally, a numerical method with an optimized step size (\(\Delta\theta=1e-3\) ) is proposed to solve the Jacobian matrix, providing a theoretical basis for high-precision motion control.
    \item \textbf{Excellent Performance Validation:} Experimental results demonstrate ByteWrist can achieve high-dynamic and large-angle motion. Compared with the Kinova dual-arm robot with serial wrists, ByteMini (integrated with ByteWrist) features higher integration and greater flexibility. ByteMini completes 116 hours of data collection for dexterous clothes-hanging tasks and realizes fully automated operation, verifying ByteWrist’s ability to cooperate with dual arms for high-precision deformable object manipulation.    
\end{enumerate}

For future improvements, we will consider optimize the structure parameters for extending the motion range. Meanwhile more light-weight design and reliable electrical wire routing will also be considered.

\section{ACKNOWLEDGMENT}
The authors sincerely thank Jiajun Zhang, Mingyu Lei, Yang Liu and Hao Niu for setting up the Kinova dual-arm robot. Meanwhile, special thanks go to the GR-3 Team, who provides an excellent VLA model for dexterous clothing-hanging manipulation task.

\vspace{0.5cm}

\bibliographystyle{unsrt}
\bibliography{main}

\begin{thebibliography}{10}

\bibitem{1-unstructuredsettings}
Liyana Wijayathunga, Rassau Alexander, and Chai Douglas.
\newblock Challenges and solutions for autonomous ground robot scene understanding and navigation in unstructured outdoor environments: A review.
\newblock {\em Applied Sciences}, 13.17:9877, 2023.

\bibitem{2-unstructuredsettings}
Klamt Tobias and et~al.
\newblock Flexible disaster response of tomorrow: Final presentation and evaluation of the centauro system.
\newblock {\em IEEE Robotics and Automation Magazine}, 26.4:59--72, 2019.

\bibitem{1-Traditionalserialroboticwrists}
Dragomir Nenchev, Tsumaki Yuichi, and Takahashi Mitsugu.
\newblock Singularity-consistent kinematic redundancy resolution for the srs manipulator.
\newblock In {\em International Conference on Intelligent Robots and Systems (IROS)}, volume~2. IEEE/RSJ, 2009.

\bibitem{2-Traditionalserialroboticwrists}
Paul Zsombor-Murray and Gfrerrer Anton.
\newblock 3r wrist positioning-a classical problem and its geometric background.
\newblock In {\em Computational Kinematics: Proceedings of the 5th International Workshop on Computational Kinematics}, volume~4. Springer Berlin, 2004.

\bibitem{3-Traditionalserialroboticwrists}
Fan Hangbing, Wei Guowu, and Ren Lei.
\newblock Prosthetic and robotic wrists comparing with the intelligently evolved human wrist: A review.
\newblock {\em Robotica}, 40.11:4169--4191, 2022.

\bibitem{4-Traditionalserialroboticwrists}
Xu~Feng, Zi~Bin, Yu~Zhaoyi, Zhao Jiahao, and Ding Huafeng.
\newblock Design and implementation of a 7-dof cable-driven serial spray-painting robot with motion-decoupling mechanisms.
\newblock {\em Mechanism and Machine Theory}, 192:105549, 2024.

\bibitem{1-parallelroboticswrists}
Peter Vischer and Clavel Reymond.
\newblock Argos: A novel 3-dof parallel wrist mechanism.
\newblock {\em The International Journal of Robotics Research}, 19.1:5--11, 2000.

\bibitem{2-parallelroboticswrists}
Pang Zaixiang and et~al.
\newblock Design and analysis of a flexible, elastic, and rope‐driven parallel mechanism for wrist rehabilitation.
\newblock {\em Applied Bionics and Biomechanics}, 2020.1:8841400, 2020.

\bibitem{3-parallelroboticswrists}
Wu~Guanglei and Niu Bin.
\newblock Dynamic stability of a tripod parallel robotic wrist featuring continuous end-effector rotation used for drill point grinder.
\newblock {\em Mechanism and Machine Theory}, 129:36--50, 2018.

\bibitem{4-parallelroboticswrists}
Wu~Yuanqing and Carricato Marco.
\newblock Design of a novel 3-dof serial-parallel robotic wrist: A symmetric space approach.
\newblock {\em Robotics Research}, 1:389--404, 2017.

\bibitem{5-parallelroboticswrists}
Sharafatdin Yessirkepov, Umurzakov Timur, and Folgheraiter Michele.
\newblock Design and analysis of a parallel elastic shoulder joint for humanoid robotics application.
\newblock {\em IEEE Access}, 2025.

\bibitem{6-parallelroboticswrists}
Baggetta Mario and et~al.
\newblock Virtual and physical prototyping of a cable-driven compliant robotic wrist.
\newblock {\em IEEE/ASME Transactions on Mechatronics}, 2025.

\bibitem{7-parallelroboticswrists}
Pollen Robotics.
\newblock Orbita : A 3-dof joint on reachy.
\newblock Forum, 2020.

\bibitem{8-parallelroboticswrists}
Pollen Robotics.
\newblock Reachy 2 is the first open-source humanoid robot specifically designed for the development of embodied ai and real-world applications.
\newblock Tech Report, 2024.

\bibitem{1-RPY}
Cheng Min and et~al.
\newblock Development of a redundant anthropomorphic hydraulically actuated manipulator with a roll-pitch-yaw spherical wrist.
\newblock {\em Frontiers of Mechanical Engineering}, 16.4:698--710, 2021.

\bibitem{2-RPY}
Bai Quan and et~al.
\newblock Coordinated motion planning of the mobile redundant manipulator for processing large complex components.
\newblock {\em The International Journal of Advanced Manufacturing Technology}, 121.9:6703--6721, 2022.

\bibitem{3-RPY}
Da~Song and et~al.
\newblock Modeling and control system experiment of a novel series three-axis stable platform.
\newblock {\em Mechanical Sciences}, 15.1:209--221, 2024.

\bibitem{4-RPY}
Marcelo Ang and Tourassis Vassilios.
\newblock Singularities of euler and roll-pitch-yaw representations.
\newblock {\em IEEE Transactions on Aerospace and Electronic Systems}, 3:317--324, 2007.

\bibitem{QDD}
Paul Zsombor-Murray and Gfrerrer Anton.
\newblock Mini cheetah: A platform for pushing the limits of dynamic quadruped control.
\newblock In {\em Mini cheetah: A platform for pushing the limits of dynamic quadruped control.} IEEE, 2019.

\bibitem{QDD1}
Katz B.G.
\newblock A low cost modular actuator for dynamic robots.
\newblock Doctoral dissertation, 2018.

\bibitem{QDD2}
P.M Wensing, A~Wang, S~Seok, D~Otten, J~Lang, and S~Kim.
\newblock Proprioceptive actuator design in the mit cheetah: Impact mitigation and high-bandwidth physical interaction for dynamic legged robots.
\newblock {\em IEEE Transactions on Robotics}, 33:509--522, 2017.

\bibitem{Kinova}
Campeau-Lecours Alexandre and et~al.
\newblock Kinova modular robot arms for service robotics applications.
\newblock In {\em Rapid Automation: Concepts, Methodologies, Tools, and Applications}, pages 693--719. IGI global, 2019.

\bibitem{GR3}
Cheang Chilam and et~al.
\newblock Gr-3 technical report.
\newblock arXiv preprint, 2025.

\bibitem{GR2}
Cheang Chilam and et~al.
\newblock Gr-2: A generative video-language-action model with web-scale knowledge for robot manipulation.
\newblock arXiv preprint, 2024.

\bibitem{vla}
Kawaharazuka Kento and et~al.
\newblock Vision-language-action models for robotics: A review towards real-world applications.
\newblock TechRxiv, 2025.

\bibitem{vla1}
Sapkota Ranjan and et~al.
\newblock Vision-language-action models: Concepts, progress, applications and challenges.
\newblock arXiv preprint, 2025.

\bibitem{vla2}
TianYu Xiang and et~al.
\newblock Parallels between vla model post-training and human motor learning: Progress, challenges, and trends.
\newblock arXiv preprint, 2025.

\bibitem{vla3}
Ma~Yueen and et~al.
\newblock A survey on vision-language-action models for embodied ai.
\newblock arXiv preprint, 2024.

\end{thebibliography}

\end{document}